\documentclass{article}
\usepackage[utf8]{inputenc}
\pdfoutput=1
\usepackage[sc,osf]{mathpazo}
\usepackage[height=8.5in,width=6in,letterpaper]{geometry}
\usepackage[sort&compress,numbers]{natbib}
\usepackage[colorlinks=true,citecolor=blue,breaklinks]{hyperref}
\usepackage[hyphenbreaks]{breakurl} %for arXiv's idiotic hyperref!

\makeatletter
\newlength\aftertitskip     \newlength\beforetitskip
\newlength\interauthorskip  \newlength\aftermaketitskip

\renewcommand\thefootnote{\textcolor{red}{\arabic{footnote}}}

\def\maketitle{\par
 \begingroup
   \def\thefootnote{\color{red}\fnsymbol{footnote}}
   \def\@makefnmark{\hbox to 4pt{$^{\@thefnmark}$\hss}}
   \@maketitle \@thanks
 \endgroup
\setcounter{footnote}{0}
 \let\maketitle\relax \let\@maketitle\relax
 \gdef\@thanks{}\gdef\@author{}\gdef\@title{}\let\thanks\relax}

\def\@startauthor{\noindent \normalsize\bf}
\def\@endauthor{}
\def\@starteditor{\noindent \small {\bf Editor:~}}
\def\@endeditor{\normalsize}
\def\@maketitle{\vbox{\hsize\textwidth
 \linewidth\hsize \vskip \beforetitskip
 {\begin{center} \LARGE\@title \par \end{center}} \vskip \aftertitskip
 {\def\and{\unskip\enspace{\rm and}\enspace}%
  \def\addr{\small\it}%
  \def\email{\hfill\small\tt}%
  \def\name{\normalsize\bf}%
  \def\AND{\@endauthor\rm\hss \vskip \interauthorskip \@startauthor}
  \@startauthor \@author \@endauthor}
}}

\makeatother

\pdfoutput=1                    % force pdflatex to run

\usepackage{floatrow} % For putting table and figure sude by side
\newfloatcommand{capbtabbox}{table}[][\FBwidth]

\usepackage{graphicx}
\graphicspath{{figures/}}
\usepackage{amsmath,amsthm,amssymb,bm} % define this before the line numbering.
\usepackage{amsfonts}
\usepackage{mathrsfs}
\usepackage{multirow}
\usepackage{xspace}
\usepackage{array}
\usepackage{enumerate}
\usepackage{algorithm}
\usepackage{algorithmic}
\usepackage{appendix}
\usepackage{wrapfig}
\usepackage{sidecap}
\usepackage{booktabs}
\usepackage[caption=false]{subfig}
\usepackage{booktabs}       % professional-quality tables
\usepackage{amsfonts}       % blackboard math symbols
\usepackage{nicefrac}       % compact symbols for 1/2, etc.
\usepackage{microtype}      % microtypography
\usepackage{subfig}
\usepackage{ifthen}
\usepackage[font=small,labelfont=bf]{caption}
\usepackage{booktabs}       % professional-quality tables
\usepackage{amsfonts}       % blackboard math symbols
\usepackage{nicefrac}       % compact symbols for 1/2, etc.
\usepackage{microtype}      % microtypography
\usepackage[export]{adjustbox} % vertical box adjustment
\usepackage{tabularx, booktabs, ragged2e}
\usepackage{tikz}
\usetikzlibrary{shapes.geometric,patterns,positioning,arrows}
\usepackage{rotating} 
\usepackage{pgfplots}
\usepgfplotslibrary{patchplots}
\pgfplotsset{compat=1.12}
\numberwithin{equation}{section}

\usepackage[capitalize]{cleveref}

\newcount\COMMENTs  % 0 suppresses notes to selves in text
\usepackage{color}
\definecolor{darkgreen}{rgb}{0,0.5,0}
\definecolor{purple}{rgb}{1,0,1}
\newcommand{\comm}[2]{\ifnum\COMMENTs=1\textcolor{#1}{#2}\fi}
\newcommand{\vX}{{\mathbf X}}
\newcommand{\vY}{{\mathbf Y}}
\newcommand{\vD}{{\mathbf D}}
\newcommand{\vd}{{\mathbf d}}
\newcommand{\vx}{{\mathbf x}}

\newcommand{\vy}{{\mathbf y}}
\newcommand{\vz}{{\mathbf z}}

\newcommand{\bs}[1]{\boldsymbol{\mathbf{#1}}}

\title{Unsupervised Multi-Target Domain Adaptation: An Information Theoretic Approach}

\author{\name Behnam Gholami \email{bb510@cs.rutgers.edu}\\ 
  \name Pritish Sahu \email{ps851@cs.rutgers.edu}\\
  \name Ognjen (Oggi) Rudovic \email{orudovic@mit.edu}\\
  \name Konstantinos Bousmalis \email{Konstantinos@google.com}\\
  \name Vladimir Pavlovic \email{vladimir@cs.rutgers.edu}\\
}

\begin{document}
\maketitle

\begin{abstract} 

	Unsupervised domain adaptation (\textbf{uDA}) models focus on pairwise adaptation settings where there is a single, labeled, source and a single target domain. However, in many real-world settings one seeks to adapt to multiple, but somewhat similar, target domains. Applying pairwise adaptation approaches to this setting may be suboptimal, as they fail to leverage shared information among multiple domains.  In this work we propose an information theoretic approach for domain adaptation in the novel context of multiple target domains with unlabeled instances and one source domain with labeled instances.  Our model aims to find a shared latent space common to all domains, while simultaneously accounting for the remaining private, domain-specific factors.  Disentanglement of shared and private information is accomplished using a unified information-theoretic approach, which also serves to establish a stronger link between the latent representations and the observed data.  The resulting model, accompanied by an efficient optimization algorithm, allows simultaneous adaptation from a single source to multiple target domains.
%The optimization task of the proposed model is minimax saddle point problems that can be optimized by adversarial training.
We test our approach on three challenging publicly-available datasets, showing that it outperforms several popular domain adaptation methods.
\end{abstract} 
\section{Introduction}\label{in}
%Traditional machine learning algorithms assume that the training and test data coming from the same underlying distribution~\cite{sun2015return}. 
In real-world data, the training and test data often do not come from the same underlying distribution~\cite{sun2015return}. For instance, in the task of object recognition/classification from image data, this is may be due to the image noise, changes in the object view, etc., which induce different biases in the observed data sampled during the training and test stage. Consequently, assumptions made by traditional learning algorithms are often violated, resulting in degradation of the algorithms' performance during inference of test data. Domain Adaptation (\textbf{DA}) approaches (e.g., ~\cite{fernando2013unsupervised,gong2012geodesic,kodirov2015unsupervised,yoo2016pixel}) aim to tackle this by transferring knowledge from a source domain (training data) to an unlabeled target domain (test data) to reduce the discrepancy between the source and target data distributions, typically by exploring domain-invariant data structures. %In supervised DA (see below) this is addressed using {\it labeled} training data (the source domain), and trying to transfer the expert knowledge (i.e., the labels) to the test data (the target domain) by reducing the discrepancy between the source and target data distributions. 

Existing \textbf{DA} methods can be divided into: (semi)supervised \textbf{DA}, and unsupervised \textbf{DA}~\cite{csurka2017comprehensive}. The former assume that in addition to the labeled data of the source domain, some labeled data from the target domain are also available for training/adapting the classifiers. By contrast, the latter does not require any labels from the target domain but rather explores the similarity in the data distributions of the two domains. In this work, we focus on the unsupervised DA (\textbf{uDA}) scenario, which is more challenging due to the lack of correspondences in source and target labels. 

Most works on \textbf{uDA} today focus on a single-source-single-target-domain scenario. However, in many real-world applications, unlabeled data may come from different domains, thus, with different statistical properties but with common task-related content. For instance, we may have access to images of the same class of objects (e.g., cars) recorded by various types of cameras, and/or under different camera views and at different times, rendering multiple different domains (e.g., datasets). Likewise, facial expressions of emotions, such as joy and surprise, shown by different people and recorded under different views, result in multiple domains with varying data distributions. In most cases, these domains have similar {\it underlying} data distributions, which can be leveraged to build more effective and robust classifiers for tasks such as the object or emotion recognition across multiple datasets/domains. To this end, traditional \textbf{uDA} methods focus on the single-source-single-target \textbf{DA} scenario. However, in the presence of multiple domains, as typically encountered in real-world settings, this pair-wise adaptation approach may be suboptimal as it fails to leverage simultaneously the knowledge shared across multiple task-related domains.

Recently, Zhao et al.~\cite{zhaoiclr2018} showed that by having access to multiple source domains can facilitate better adaptation to a single target domain, when compared to the pair-wise \textbf{DA} approach. While this is intuitive due to the access to multiple {\it \textbf{labelled}} source domains, offering more adaptation flexibility for the target domain (i.e., by efficiently exploring the data labels across multiple source domains that are most related to the target domain), it comes at the expense of the data labelling in multiple source domains, which can be costly and time-consuming.

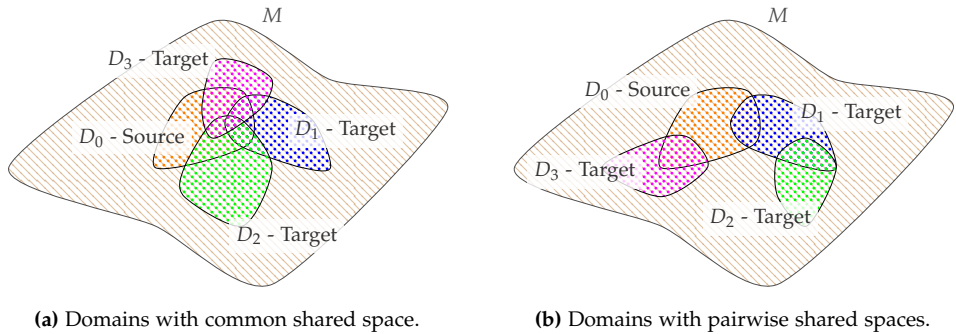
\begin{figure}[tb]
     \centering
     \begin{tabular}{cc}
     \subfloat[Domains with common shared space. \label{fig:manifold_shared}]{
     	\resizebox{.4\textwidth}{!}{\begin{tikzpicture}

    % Manifold
    \draw[smooth cycle, tension=0.4, fill=white, pattern color=brown, pattern=north west lines, opacity=0.7] plot coordinates{(2,2.5) (-1.5,0) (1,-1) (3,-2) (6,1) (4,1.5)} node at (3,2.6) {$M$};

    % Help lines
    %\draw[help lines] (-3,-6) grid (8,6);

    % Subsets
    \draw[smooth cycle, pattern color=orange, pattern=crosshatch dots]
        plot coordinates {(1,0) (1.5, 1.2) (2.5,1.3) (2.6, 0.4)}
        node [label={[label distance=-0.3cm, xshift=-2cm, fill=white, opacity=0.8]:$D_0$ - Source}] {};

    \draw[smooth cycle, pattern color=blue, pattern=crosshatch dots]
        plot coordinates {(4, 0) (3.7, 0.8) (3.0, 1.2) (2.5, 1.2) (2.2, 0.8) (2.3, 0.5) (2.6, 0.3) (3.5, 0.0)}
        node [label={[label distance=-0.8cm, xshift=.75cm, yshift=1cm, fill=white, opacity=0.8]:$D_1$ - Target}] {};

    \draw[smooth cycle, pattern color=green, pattern=crosshatch dots]
            plot coordinates {(3, 0) (2.5, 0.8) (2.0, 0.8) (1.5, 0) (1.5, -.5) (2.5, -1) }
            node [label={[label distance=-0.8cm, xshift=.75cm, yshift=.2cm, fill=white, opacity=0.8]:$D_2$ - Target}] {};

    \draw[smooth cycle, pattern color=magenta, pattern=crosshatch dots]
                    plot coordinates {(3, 1.5) (2.5, 1.8) (2.0, 1.8) (1.8, 1) (2, .5) (2.8, 1) }
                    node [label={[label distance=-0.8cm, xshift=-1.75cm, yshift=1.2cm, fill=white, opacity=0.8]:$D_3$ - Target}] {};

\end{tikzpicture}}
     } &
     \subfloat[Domains with pairwise shared spaces. \label{fig:manifold_linked}]{
        \resizebox{.4\textwidth}{!}{\begin{tikzpicture}

    % Manifold
    \draw[smooth cycle, tension=0.4, fill=white, pattern color=brown, pattern=north west lines, opacity=0.7] plot coordinates{(2,2.5) (-1.5,0) (1,-1) (3,-2) (6,1) (4,1.5)} node at (3,2.6) {$M$};

    % Help lines
    %\draw[help lines] (-3,-6) grid (8,6);

    % Subsets
    \draw[smooth cycle, pattern color=orange, pattern=crosshatch dots]
        plot coordinates {(1,0) (1.5, 1.2) (2.5,1.3) (2.6, 0.4)}
        node [label={[label distance=-0.3cm, xshift=-2cm, yshift=.8cm, fill=white, opacity=0.8]:$D_0$ - Source}] {};

    \draw[smooth cycle, pattern color=blue, pattern=crosshatch dots]
        plot coordinates {(4, 0) (3.7, 0.8) (3.0, 1.2) (2.5, 1.2) (2.2, 0.8) (2.3, 0.5) (2.6, 0.3) (3.5, 0.0)}
        node [label={[label distance=-0.8cm, xshift=.75cm, yshift=1.3cm, fill=white, opacity=0.8]:$D_1$ - Target}] {};

    \draw[smooth cycle, pattern color=green, pattern=crosshatch dots]
            plot coordinates {(4, 0) (3.5, 0.5) (3.0, 0.2)  (3, -.5) (3.5, -1) }
            node [label={[label distance=-0.8cm, xshift=-.8cm, yshift=.5cm, fill=white, opacity=0.8]:$D_2$ - Target}] {};

    \draw[smooth cycle, pattern color=magenta, pattern=crosshatch dots]
                    plot coordinates {(1.5, 0.5) (1.0, 0.5) (0, 0) (1, -.5) (1.8, 0) }
                    node [label={[label distance=-0.8cm, xshift=-2.1cm, yshift=0.3cm, fill=white, opacity=0.8]:$D_3$ - Target}] {};

\end{tikzpicture}}
     }
     \end{tabular}
     \caption{Illustration of domains with common (a) and pairwise-shared spaces (b).  We tackle the domain adaptation task when all domains share a common task/space, which is then leveraged to transfer knowledge across multiple target domains.}
     \label{fig:manifolds}
\end{figure}

In either case, a single source domain or readily available multiple source domains, to the best of our knowledge, a simultaneous adaptation to {\it \textbf{multiple}} and {\it \textbf{unlabelled}} target domains remains an unexplored DA scenario. However, this \textbf{DA} scenario is important as we usually have access to multiple unlabeled domains; yet, the adaptation process is also more challenging due to the lack of supervision in the target domains. Still, multi-target \textbf{DA} can have advantages over a single-target \textbf{DA} when: (i) there is direct knowledge sharing between the source and multiple target domains (Fig.~\ref{fig:manifold_shared}), and (ii) the source and a target domain are related through another target domain (Fig.~\ref{fig:manifold_linked}). While this seems intuitive, it is critical how the data from multiple {\it \textbf{unlabelled}} target domains are leveraged within the multi-target \textbf{DA} approach, in order to improve its performance over the single target \textbf{DA} approaches and naive fusion of multiple target domains.\\
To this end, we propose a Multi-Target DA-Information-Theoretic-Approach (\textbf{MTDA-ITA}) for single-source-multi-target \textbf{DA}. We exploit a single source domain and focus on multiple target domains to investigate the effects of multi-target \textbf{DA}; however, the proposed model can easily be extended to multiple source domains. This approach leverages the data from
multiple target domains to improve performance compared to individually learning from pairwise source-target domains. Specifically, we simultaneously factorize the information from each available target domain and learn separate subspaces for modeling the shared (i.e., correlated across the domains) and private (i.e., independent between the domains) subspaces of the data~\cite{salzmann2010factorized}. To this end, we employ deep learning to derive an information theoretic approach where we jointly maximize the mutual information between the domain labels and private (domain-specific) features, while minimizing the mutual information between the the domain labels and the shared (domain-invariant) features. Consequently, the more robust feature representations are learned for each target domain by exploiting dependencies between multiple target domains. We show on benchmark datasets for \textbf{DA} that this approach leads to overall improved performance on each target domain, compared to independent \textbf{DA} for each pair of source-target domains, or the naive combination of multiple target domains, and state-of-the-art models applicable to the target task.
\section{Preliminaries}
\subsection{Information Theory: Background}\label{sec:back}
Let $\vx = (x_1, x_2, ..., x_n)$ denote a $n$-dimensional random variable with probability density function (pdf) given by $p(\vx)$. Shannon differential entropy is defined in the usual way as $H(\vx) = -\mathbb{E}_\vx\left[\ln p(\vx)\right]$ where $\mathbb{E}$ denotes the expectation operator. Let $\vz = (z_1, z_2, ..., z_m)$ denote a $m$-dimensional random variable with pdf $p(\vz)$. Then mutual information between two random variables, $\vx$ and $\vz$, is defined as $I(\vx;\vz)=H(\vx)+H(\vz)-H(\vx,\vz)$. Mutual information can also be viewed as the reduction in uncertainty about one variable given another variable---i.e.,  $I(\vx;\vz)=H(\vx)-H(\vx|\vz)=H(\vz)-H(\vz|\vx)$.
\section{Method}\label{pm}
In this section, we describe our proposed information theoretic approach that supports  domain  adaptation for multiple target domains simultaneously, finding factorized latent spaces that are non-redundant, and that can capture explicitly the shared (domain invariant) and the private (domain dependent) features of the data well suited for better generalization  for domain  adaptation.
\subsection{Problem Formulation}
Without loss of generalizability, we consider a multi-class ($K$-class) classification problem as the running example. Furthermore, let $(\vX, \vY, \vD) = \{({\vx}_i, {\vy}_i, \vd_i)\}_{i=0}^{N}$ be a collection of $M$ domains (a labeled source domain, and $M-1$ unlabeled target domains), where ${\vx_i}$ denotes the $i$-th sample, and $\vy_i = [y_i^1,y_i^2,...,y_i^K]$ and ${\vd_i=[d_i^1,d_i^2,...,d_i^M]}$ are the $K$-D and $M$-D encoding of the class and domain labels for $\vx_i$, respectively. Note that the class labels are only available for the source samples.
 
The latent space representation of the data point $\vx$ is denoted as $\bs z = [\bs z_s,\bs z_p]$, where $\bs z_s$ and $\bs z_p$ are the (latent) shared and private features of the data point $\bs x$, respectively. Let $\bs z_s$ and $\bs z_p$ be some stochastic function of ($\vx$, $\bs d$) parameterized by ($\theta_s$, $\theta_p$), respectively, and $\bs y$ be some stochastic function of $\bs z_s$ parameterized by $\theta_c$. 
%Then we seek to model a joint distribution given by:
%\begin{equation}
%p(\vx, \vy, \bs d,\bs z_s, \bs z_p;\theta_s,\theta_p,\theta_c) = p(\vx)p(\bs d)p(\bs z_s|\vx;\theta_s)p(\bs z_p|\vx;\theta_p)p(\vy|\vz_s;\theta_c),
%\end{equation}
%where $p(\vx)$ and $p(\bs d)$ denote the underlying (true) data distribution and domain label distribution, respectively. We use the above joint distribution to 
We propose to maximize the following objective function:
\begin{equation} \label{eq:loss}
\begin{split}
\mathcal{L}(\theta_s,\theta_p,\theta_c;\vx,\vy, \bs d) &=  \lambda_r I(\vx;\bs z) + \lambda_c I(\bs y;\bs z_s) + \lambda_d \big(I(\bs d;\bs z_p)- I(\bs d;\bs z_s)\big),
\end{split}
\end{equation}
where $I(x;y)$ denotes the Mutual Information between the random variables $\bs x$ and $\bs y$. $\lambda_r,\lambda_c$ and $\lambda_d$ denote the hyper-parameters controlling the weights of the objective terms.
The proposed objective function~(\ref{eq:loss}) maximizes the three terms described below:
\begin{itemize}
\item $I(\vx;\bs z):$ encourages the latent features (both shared and private) to preserve information about the data samples (that can be used to reconstruct $\vx$ from $\vz$).

\item $I(\bs y;\bs z_s)$: enables to correctly predict the true class label of the samples out of their common shared features.  
%\item $I(\bs d;\bs z_p):$ preserves information about the domain label, rendering private (domain-specific) features.
 
\item $I(\bs d;\bs z_p) - I(\bs d;\bs z_s):$ encourages the latent private features to preserve the information about the domain label and penalizes the latent shared features to be domain informative. This not only reduces the redundancy in the shared and private features, but also, penalizes the redundancy of different private spaces, while preserving the shared information. 
\end{itemize}
An additional term could be used to minimize the mutual information between the shared ($\bs z_s$) and private ($\bs z_p$) features. However, computing the mutual information (even approximating it) is intractable due to the highly complex joint distribution $p(\bs z_s,\bs z_p)$. Since we want $\bs z_s$ and $\bs z_p$ features to encode different aspects of $\vx$, we enforce such constraint by jointly maximizing the term: $I(\bs d;\bs z_p)-I(\bs d;\bs z_s)$. 

\subsection{Optimization}
The following lower bound for mutual information is derived using the non-negativity of KL-divergence~\cite{barber2003algorithm}; i.e., $\Sigma_{\vx}p(\vx|\vz)\ln\frac{p(\vx|\vz)}{q(\vx|\vz)} \ge 0$ gives:
\begin{equation}\label{eq:xi_z}
I(\vx;\vz) \ge H(\vx) + \mathbb{E}_{p(\vx, \vz)}[\ln q(\vx | \vz;\phi)],
\end{equation}
where $H(\bs x)$ denotes the Shanon Entropy of the random variable $\bs x$. $q(\vx | \vz;\phi)$ is any arbitrary distribution parametrized by $\phi$. We need a variational distribution $q(\vx | \vz;\phi)$ because the posterior distribution $p(\vx|\vz)=p(\vz|\vx)p(\vx)/p(\vz)$ is intractable since the true data distribution $p(\vx)$ is assumed to be unknown. Similarly, we can derive lower bounds for $I(\bs d;\bs z_p) \ge H(\bs d) + \mathbb{E}_{p(\bs d, \bs z_p)}[\ln q(\bs d | \bs z_p;\psi)]$ and $I_{}(\bs d;\bs z_s) \ge H(\bs d) + \mathbb{E}_{p(\bs d, \bs z_s)}[\ln q_{}(\bs d | \bs z_s;\psi)]$, where $q(\bs d | \bs z_p;\psi)$ is any arbitrary distribution parametrized by $\psi$.\footnote{Note that, for simplicity, we shared the parameters $\psi$ between the approximate posterior distributions $q(\bs d | \bs z_s,\psi)$ and $q(\bs d | \bs z_p;\psi)$.}
We further compute $I(\bs y;\bs z_s)$ as $I(\bs y;\bs z_s) = H(\bs y) + \mathbb{E}_{p(\bs y, \bs z_s)}[\ln p(\bs y | \bs z_s)]$. 

\begin{figure}[tb]
     \centering
     \resizebox{0.5\textwidth}{!}{\input{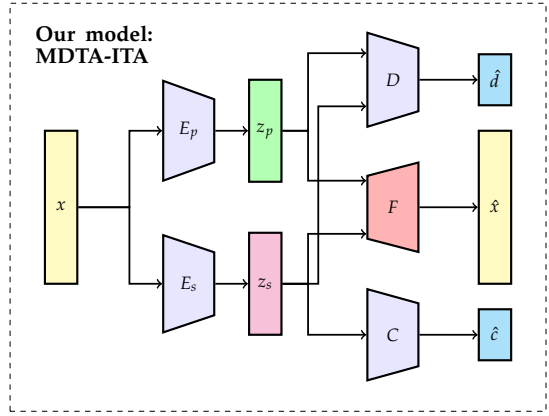}}
     \caption{\small MDTA-ITA: The encoder $E_{s}(\bs x)$ captures the feature representations ($z_s$) for a given input sample $\bs x$ that are shared among domains. $E_p(\bs x)$ captures domain--specific private features ($z_p$) using the {\it shared} private encoder. The shared decoder $F(z_p,z_s)$ learns to reconstruct the input sample by using both the private and shared features. The domain classifier $D$ learns to correctly predict the domain labels of the actual samples from both their shared and private features while the classifier $C$ learns to correctly predict the class labels from the shared features. %Contrast this to a generalization of DSN to multiple target domains, where each target domain $i$ has its own encoder $E_{pi}(\bs x)$. There is no discriminator $D$ to enforce separation of spaces $z_s,z_{p1},z_{p2}$.%The overview of the proposed model, MDTA-ITA, (left) and comparison to a generalization of DSN~\cite{bousmalis2016domain} to two target domains (right).  In MDTA-ITA, the encoder $E_{s}(\bs x)$ learns to capture representation components for a given input sample $\bs x$ that are shared among domains. The private encoder $E_p(\bs x)$ (also shared among all domains) learns to capture domain--specific components of the representation. The shared decoder $F(z_p,z_s)$ learns to reconstruct the input sample by using both the private and shared representations. The domain classifier $D$ learns to correctly predict the domain labels of the actual samples from their both shared and private representations. The classifier $C$ learns to correctly predict the class labels of the samples from their shared representations. Contrast this to a generalization of DSN to multiple (here two) target domains, where each target domain $i$ has its own encoder $E_{pi}(\bs x)$. There is no discriminator $D$ to enforce separation of spaces $z_s,z_{p1},z_{p2}$.
%The overview of the proposed model, MDTA-ITA. The encoder $E_{s}(\bs x)$ learns to capture representation components for a given input sample $\bs x$ that are shared among domains. The private encoder $E_p(\bs x)$ (also shared among all domains) learns to capture domain--specific components of the representation. The shared decoder $F(z_p,z_s)$ learns to reconstruct the input sample by using both the private and shared representations. The domain classifier $D$ learns to correctly predict the domain labels of the actual samples from both their shared and private representations. The classifier $C$ learns to correctly predict the class labels from the shared features. Contrast this to a generalization of DSN to multiple target domains, where each target domain $i$ has its own encoder $E_{pi}(\bs x)$. There is no discriminator $D$ to enforce separation of spaces $z_s,z_{p1},z_{p2}$.
     }
     \label{model:pm}
\end{figure}

Let next $E_s(\bs x; \theta_s)$ be a function parameterized by $ \theta_s$ that maps a sample $\bs x $ to the \textit{shared} features $\bs z_s$, and $E_p(\bs x; \theta_p)$ be
an analogous function which maps $\bs x$ to $\bs z_p$, the features that are \textit{private} to each domain (Fig.~\ref{model:pm}). We also define $F(\bs z_s, \bs z_p;\phi)$ as a decoding function mapping the concatenation of the latent features $\bs z_s$ and $\bs z_p$ to a sample reconstruction $\hat{\bs x}$, and $D(\bs z;\psi)$ as a decoding function mapping $\bs z_s$ and $\bs z_p$ to a $M$-dimensional probability vector: the predictions of the domain label $\hat{\bs d}$. Finally, $C(\bs z_s; \theta_c)$
is a task-specific function mapping $\bs z_s$ to a $K$-dimensional probability vector of the class label $\hat{\bs y}$.

By representing $p(\bs d),p(\vx), p(\vy)$ as the empirical distribution of a finite training set (e.g. $p(\bs d) = \frac{1}{N}\sum_{i=1}^N\delta(d-d_i)$) as in the case of variational autoencoders (VAE)~\cite{abbasnejad2017infinite,pu2017adversarial}, parametarizing $p(\bs z_s| \bs x)$, and $p(\bs z_p| \bs x)$ as deterministic networks $E_s(\bs x)$ and $E_p(\bs x)$ respectively, and modeling the variational distributions as $\ln q(\bs x|\bs z;\psi) = \|\bs x - F(\bs z;\phi)\|_1, \ln q(\bs d|\bs z)= \bs d^{\top} \ln D(\bs z;\psi)$, and $\ln p(\bs y|\bs z_{s})= \bs y^{\top} \ln C(\bs z_{s};\theta_c)$, where $\|.\|_1$ denotes the $L_1$ norm, the optimization task can be posed as a minimax saddle point problem, where we use adversarial training to maximize (\ref{eq:loss}) w.r.t. the stochastic parameters ($\theta_s,\theta_p,\theta_c$), and to minimize (\ref{eq:loss}) w.r.t. the variational parameters ($\phi$,$\psi$), using Stochastic Gradient Descent (SGD). 
 \subsubsection*{Optimizing the parameters $\phi$ \textbf{of the decoder} $F$}
\begin{align}\label{psi}
\hat{\phi} = \underset{\phi}{\arg \min}\; \mathcal{L}_F = 
 \frac{\lambda_r}{N}\sum_{i=1}^N\|\bs x_i - F\big(E_s(\bs x_i),E_p(\bs x_i)\big)\|_1.
 \end{align}
The decoder $F(\bs z_s,\bs z_p;\phi)$ is trained in such a way so as to minimize the difference between original input ${\bs x}$ and its decoding from corresponding shared and private features via the decoder $F(\cdot)$. 

\subsubsection*{Optimizing the parameters $\psi$ \textbf{of the domain classifier} $D$}
\begin{align}\label{phi}
\hat{\psi} = \underset{\psi}{\arg \min}\;\mathcal{L}_D=   -\frac{\lambda_d}{N}\sum_{i=1}^{N} \bs d_i^{\top} \ln D\big(E_s(\bs x_i)\big)  -\frac{\lambda_d}{N}\sum_{i=1}^{N} \bs d_i^{\top}\ln D\big(E_p(\bs x_i)\big).
\end{align}
 $D(\bs z;\psi)$ can be considered as a classifier whose task is to distinguish between the shared/private features of the different domains. More precisely, the two terms in Eq.~\ref{phi} encourage $D$ to correctly predict the domain labels from the shared and private features, respectively.

\subsubsection*{Optimizing the parameters $\theta_c$ \textbf{of the label classifier} $C$}
\begin{align}
\hat{\theta}_c = \underset{\theta_c}{\arg \min}\; 
\big\{- H(\bs y) - \mathbb{E}_{p(\bs y, \bs z_s)}\big[\ln p(\bs y | \bs z_s)\big]\big\}.
\end{align}
Since we have access to the source labels, $H(\bs y)$ is a constant for source samples. 
%and $ \mathbb{E}_{p(\bs y, \bs z_s)}[\ln p(\bs y | \bs z_s)]$ for the source samples can be approximated as:
%\begin{equation}
 %\mathbb{E}_{p(\bs y, \bs z_s)}\big[\ln p(\bs y | \bs z_s)\big] \approx %\frac{1}{N_s}\sum_{i=1}^{N_s} \vy_i^{\top}\ln C\big( E_s(\bs x_i) \big),
%\end{equation}
we can approximate $H[\vy]$ for the target samples using the output of the classifier $C$, leading to the following optimization problem:
\begin{align}\label{theta-c}
\hat{\theta_c} = \underset{\theta_c}{\arg \min}\;\mathcal{L}_C=& 
- \frac{1}{N}\sum_{i=1}^{N_s} \bs y_i^{T}\ln C\big( E_s(\bs x_i) \big) - \frac{\lambda_c}{N-N_s}\sum_{i=N_s+1}^{N} C\big( E_s(\bs x_i) \big)^{\top}\ln C\big( E_s(\bs x_i) \big)\nonumber \\
& + \frac{\lambda_c}{N-N_s}\sum_{i=N_s+1}^{N}C\big( E_s(\bs x_i) \big)^{\top}\ln\bigg(\frac{1}{N-N_s}\sum_{i=N_s+1}^{N}C\big( E_s(\bs x_i) \big)\bigg), 
\end{align}
where $N_s$ denotes the number of source samples. Intuitively, we enforce the classifier $C(\bs z_s;\theta_c)$ to correctly predict the class labels of the source samples by the first term in Eq.~\ref{theta-c}. We use the second term to minimize the entropy of $p(\bs y|\vz_s)$ for the target samples; effectively, reducing the effects of "confusing" labels of target samples, as given by $p(\bs y|\vz_s)$ that leads to decision boundaries occur far away from target data-dense regions in the feature space.  The intuition behind the last term is that if we minimize the entropy only(second term), we may arrive at a degenerate solution where every point $x_t$ is assigned to the same class. Hence, the last term encourages the classifier $C(\cdot)$ to have balanced labeling for the target samples where it reaches its minimum, $\ln K$, when each class is selected with uniform probability. 

\subsubsection*{Optimizing the parameter $\theta_p$ \textbf{of the private encoder} $E_p$}
\begin{align}\label{theta-p}
\hat{\theta_p} = \underset{\theta_p}{\arg \min}\;\mathcal{L}_P= &\frac{\lambda_r}{N}\sum_{i=1}^N \|\bs x_i - F\big(E_s(\bs x_i),E_p(\bs x_i)\big)\|_1 -\frac{\lambda_d}{N}\sum_{i=1}^N \bs d_i^{\top}D\big(E_p(\bs x_i)\big). 
\end{align}
The first term in Eq.~\ref{theta-p} encourages the private encoder $E_p(\bs x;\theta_p)$ to preserve the recovery ability of the private features. The second term, $E_p(\cdot)$, enforces distinct private features be produced for each domain by penalizing the representation redundancy in different private spaces. This, in turn, encourages moving this common information from multiple domains to their shared space.

\subsubsection*{Optimizing the parameter $\theta_s$ \textbf{of the shared encoder} $E_s$}
\begin{align}\label{theta-s}
\hat{\theta_s} = \underset{\theta_s}{\arg \min}\;\mathcal{L}_S= &\frac{\lambda_r}{N}\sum_{i=1}^N \|\bs x - F\big(E_s(\bs x_i),E_p(\bs x_i)\big)\|_1  \nonumber - \frac{\lambda_c}{N}\sum_{i=1}^{N_s} \bs y_i^{T}\ln C\big( E_s(\bs x_i) \big)\\
&-\frac{\lambda_d}{N}\sum_{i=1}^N \bs d_i^{\top}\ln D\big(E_s(\bs x_i)\big)- \frac{\lambda_c}{N-N_s}\sum_{i=N_s+1}^{N} C\big( E_s(\bs x_i) \big)^{\top}\ln C\big( E_s(\bs x_i) \big)\nonumber \\
& + \frac{\lambda_c}{N-N_s}\sum_{i=N_s+1}^{N}C\big( E_s(\bs x_i) \big)^{\top}\ln\bigg(\frac{1}{N-N_s}\sum_{i=N_s+1}^{N}C\big( E_s(\bs x_i) \big)\bigg).
\end{align}
The first term in Eq.~\ref{theta-s} encourages the shared encoder $E_s(\bs x;\theta_s)$ to preserve the recovery ability of the shared features. The second term is the source domain classification loss penalty that encourages $E_s$ to produce discriminative features for the labeled source samples. The third term simulates the adversarial training by trying to fool the domain classifier $D(\cdot)$ when predicting the domain labels $\bs d$, given the shared features $\bs z_s$. The effect of this is two-fold: 
(i) the rendered shared features are more distinct from the corresponding private features, (ii) the shared features of different domains are encouraged to be similar to each other. The last two terms encourage $E_s(\cdot)$ to produce the shared features for target samples so that the classifier is confident on the unlabeled
target data, driving the shared features away from the decision boundaries. To train our model, we alternate between updating the shared encoder $E_c(\cdot)$, the private encoder $E_p(\cdot)$, the decoder $F(\cdot)$, the classifier $C(\cdot)$, and the domain classifier $D(\cdot)$ using the SGD algorithm (see Algorithm~\ref{alg1} for more details). 
\begin{algorithm}[t]
\caption{MDTA-ITA Algorithm}
\label{alg1}
\begin{algorithmic}[1]
\REQUIRE

$\{\mathbf X, \mathbf Y, \mathbf{D}\}$:M domain datasets.\\
$\ \ \ \ \ \ \ \ \ \lambda_r,\lambda_c,\lambda_d$: Model hyper-parameters.\\
$\ \ \ \ \ \ \ \ \ \eta$: Learning rate.\\
\ENSURE

$\theta_s, \theta_p,\theta_c,\phi,\psi$: Model parameters.\\
\STATE Initialize $\theta_s, \theta_p,\theta_c,\phi,\psi$;
\REPEAT
\STATE Sample a mini-batch from each of source/target domain datasets.
\STATE Update $\{\theta_s\}$ by minimizing $\mathcal{L}_s$ in Eq.(\ref{theta-s}) through the gradient descent:
$\theta_s = \theta_s - \eta\frac{\partial \mathcal{L}_s}{\partial \theta_s}$.
\STATE Update $\{\theta_p\}$ by minimizing $\mathcal{L}_p$ in Eq.(\ref{theta-p}) through the gradient descent:$\theta_p = \theta_p - \eta\frac{\partial \mathcal{L}_p}{\partial \theta_p}$.

\STATE Update $\{\theta_c\}$ by minimizing $\mathcal{L}_c$ in Eq.(\ref{theta-c}) through the gradient descent:$\theta_c = \theta_c - \eta\frac{\partial \mathcal{L}_c}{\partial \theta_c}$.

\STATE Update $\{\phi\}$ by minimizing $\mathcal{L}_{\phi}$ in Eq.(\ref{phi}) through the gradient descent:$\phi = \phi - \eta\frac{\partial \mathcal{L}_{\phi}}{\partial \phi}$.

\STATE Update $\{\psi\}$ by minimizing $\mathcal{L}_{\psi}$ in Eq.(\ref{psi}) through the gradient descent:$\psi = \psi - \eta\frac{\partial \mathcal{L}_s}{\partial \psi}$.

\UNTIL{Convergence};
\STATE return $\{\theta_s, \theta_p,\theta_c,\phi,\psi\}$. 
\end{algorithmic}
\end{algorithm}
\section{Related Work}\label{rw}
There has been extensive prior work on domain adaptation~\cite{csurka2017comprehensive}.
%(~\cite{li2016revisiting,morerio2018minimalentropy,sener2016learning,french2018selfensembling, jiang2007instance,foster2010discriminative,chen2011co,saenko2010adapting,pan2011domain,gopalan2011domain,gong2012geodesic,long2013transfer,yang2007adapting,aytar2011tabula,jiang2008cross}). 
Recent  papers  have  focused  on transferring  deep  neural  network  representations  from  a labeled source dataset to an unlabeled target domain, where the  main strategy 
is to find a feature space such that the confusion between source and target distributions in that space is maximized (~\cite{rebuffi2017learning,benaim2017one,courty2017joint,motiian2017few,saito2017asymmetric,Zhang_2017_CVPR,Yan_2017_CVPR,bousmalis2017unsupervised}). For this, it is critical to first define a measure of divergence between source and target distributions. For instance, several methods have used the Maximum Mean Discrepancy (\textbf{MMD}) loss for this purpose (e.g., ~\cite{bousmalis2017unsupervised,zellinger2017central,long2014transfer}). \textbf{MMD} computes the norm of the difference between two domain means in the reproducing Kernel Hilbert Space (RKHS) induced by a pre-specified kernel. The Deep Adaptation Network (DAN)~\cite{long2015learning} applied MMD to layers embedded in a RKHS, effectively matching higher order statistics of the two distributions. The deep Correlation Alignment (\textbf{CORAL}) method~\cite{sun2016deep} attempts to match the mean and covariance of the two distributions. Deep Transfer Network (\textbf{DTN})~\cite{zhang2015deep} achieved source/target distribution alignment via two types of network layers based on \textbf{MMD} distance: the shared feature extraction layer, which learns a subspace that matches the marginal distributions of the source and the target samples, and the discrimination layer, which matches the conditional distributions by classifier transduction.

Recently proposed unsupervised \textbf{DA} methods (~\cite{rebuffi2017learning,benaim2017one,courty2017joint,motiian2017few,saito2017asymmetric,Zhang_2017_CVPR}) operate by training  deep  neural  networks  using adversarial training, which allows the learning of feature representations that are simultaneously discriminative of source labels, and indistinguishable between the source and target domain. For instance, Ganin et al.~\cite{ganin2014unsupervised} proposed a DA mechanism called Domain-Adversarial Training of Neural Networks (\textbf{DANN}), which enables the network to learn domain invariant representations in an adversarial way by adding a domain classifier and back-propagating inverse gradients. Adversarial Discriminative  Domain  Adaptation (\textbf{ADDA})~\cite{tzeng2017adversarial} learns a discriminative feature subspace using the source labels, followed by a separate encoding of the target data to this subspace using an asymmetric mapping learned through a domain-adversarial loss. Liu et al.~\cite{liu2017unsupervised} makes a shared-latent space assumption and proposes an unsupervised image-to-image translation (\textbf{UNIT}) framework based on Coupled GANs~\cite{NIPS2016_6544}. Another example is the pixel-level domain adaptation models that perform the distribution alignment not in the feature space but directly in raw pixel space. \textbf{PixelDA}~\cite{bousmalis2017unsupervised} uses adversarial approaches to adapt source-domain images as if drawn from the target domain while maintaining the original content.

%%% oggi

While these approaches have shown success in \textbf{DA} tasks with single source-target domains, they are not designed to leverage information from multiple domains simultaneously. More recently, Zhao et al.~\cite{zhaoiclr2018} introduced an adversarial framework called \textbf{MDAN} for multiple source single target domain adaptation where a domain classifier, induced by minimizing the H-divergence between multiple source and a target domain, is used to align their feature distributions in a shared space. Instead, in our approach we focus on multi-target \textbf{DA} where we perform adaptation of multiple {\it unlabelled} target domains. Although both our model and \textbf{MDAN} use the similar notion of the domain classifier to minimize the domain mismatch in shared space, the domain classifier induced by our information-theoretic (IT) loss also acts to separate domains in the private space (see Eqs.~\ref{phi}, and~\ref{theta-p} for more details), improving the essential reconstruction ability, similar to~\cite{bousmalis2016domain}. 

%From the probabilistic point of view, our model is VAE-inspired~\cite{bowman2015generating}. However, we propose to tackle the task using our IT approach instead of the traditional evidence lower bound (ELBO) optimization. One of the main drawbacks of ELBO-based approaches is that they can result in poor latent representation, when a powerful decoder effectively ignores the latent space~\cite{bowman2015generating}. In the spirit of recent works on improved representational learning~\cite{alemi2018fixing}, we seek to recover a good latent representation by replacing the ELBO objective with an IT-driven loss. In contrast to the unsupervised representation learning approaches, our setting allows us to further improve the latent representation using the labeled data in the source domain while leveraging the sharing of dependencies across different target domains.
\subsection{Connection to Information Theoretic Representation Learning}
The  idea  of  using  information  theoretic (IT)  objectives  for  representation learning  was  originally introduced in
Tishby $\&$ Zaslavsky~\cite{tishby2015deep}. Since their
approach for optimizing the IT objective functions relied on the iterative Blahut Arimoto algorithm~\cite{tishby2015deep}, it is not feasible to apply to DNN frameworks. 
Similar to our approach, there have been some recent works~\cite{mohamed2015variational,chalk2016relevant,alemi2018uncertainty,alemi2016deep,alemi2018fixing} to approximate the MI by applying variational bounds on \textbf{MI}, though not in the context of domain adaptation.

Mohamed $\&$ Rezende~\cite{mohamed2015variational} utilized the variational bounds on \textbf{MI}, and apply it to deep neural networks in the context of reinforcement learning. Chalk et al.~\cite{chalk2016relevant} and Amini et al.~\cite{alemi2016deep}, developed the same variational lower bound In the context of Information Bottleneck (IB) principle~\cite{tishby2015deep}, where the former applied it to sparse coding problems, and
used the kernel trick to achieve nonlinear mappings, whereas the latter applied it to deep neural networks to handle large datasets thanks to the SGD algorithm.  Achille $\&$ Soatto~\cite{achille2018information} proposed a variational bound on
the IM in the context of IB, from the perspective of variational dropout and demonstrated
its utility in learning disentangled representations for variational autoencoders.

The main difference between our method and the above methods is that these methods throw away the information in the data not related to the task by minimizing the mutual information between the data points and the latent representations that may lead to ignoring the individual characteristics (private features) of the datasets in a multiple dataset regime, whereas our method explicitly models what is unique to each domain (dataset) that improves the model’s ability to extract domain–invariant features.

In the unsupervised representation learning literature, our work is also related to the VAE-based models~\cite{bowman2015generating}. However, we propose to tackle the task using our IT approach using deterministic mappings instead of the traditional evidence lower bound (\textbf{ELBO}) optimization with stochastic mappings. One of the main drawbacks of \textbf{ELBO}-based approaches is that they can result in poor latent representation, when a powerful decoder effectively ignores the latent space~\cite{bowman2015generating}. In the spirit of recent works on improved representational learning~\cite{alemi2018fixing}, we seek to recover a good latent representation by replacing the \textbf{ELBO} objective with an IT-driven loss. In contrast to the unsupervised representation learning approaches, our setting also allows us to further improve the latent representation using the labeled data in the source domain while leveraging the sharing of dependencies across different target domains.
\subsection{Connection to Multiple Domain Transfer Networks}
Recent studies have shown remarkable success in multiple domain transfer (MDT)~\cite{choi2017stargan,anoosheh2017combogan,kameoka2018stargan,hao2018mixgan} though not in the context of the image classification, rather in the context of image generation. choi et al.~\cite{choi2017stargan} proposed \textbf{StarGAN}, a generative adversarial network capable of learning mappings among multiple domains in the contest of image to image translation framework. The goal of \textbf{StarGAN} is to train a single generator $G$ though
this requires passing in a vector along with each input to the
generator specifying the output domain desired, that learns mappings among multiple domains. To achieve this, $G$ is trained to translate an input image $\bs x$  into an output image $\bs x'$ conditioned on the target domain label $\bs d$, $G(x, d) \rightarrow x'$. Similar to our domain classifier module $D$, they introduce an auxiliary classifier that allows a single discriminator to control multiple domains.

Anoosheh et al.~\cite{anoosheh2017combogan} introduced \textbf{ComboGAN}, which decouples the domains and networks from each other. Similar to our encoder/decoder modules, \textbf{ComboGAN}'s generator networks contain encoder/decoders assigning each encoder and decoder to a domain.
They combine the encoders and decoders of the trained model like building blocks, taking as input any domain and outputting any other. For example during inference, to transform an image $\bs x$ from an arbitrary domain $\bs X$ to $\bs x'$ from domain $\bs X'$, they simply perform $x' = G_{\bs X' \bs X}(\bs x) = Decoder_X'(Encoder_X(\bs x))$. The result of $Encoder_X(\bs x)$ can even be cached when translating to other domains as not to repeat computation.

The main differences between the MDT methods and ours is that, unlike our method which does domain alignment in feature space, MDT methods adapt representations not in
feature space but rather in raw pixel space; translating samples from one domain to the “style” of a other domains.
This works well for limited domain shifts where the domains are similar in pixel-space, but can be too limiting for
settings with larger domain shifts that
results in poor performance in significant structural change of the samples in different domains. 
\subsection{Connection to Domain Separation Networks}\label{conn}
The method closest to our work is Domain Separation Networks (\textbf{DSN})~\cite{bousmalis2016domain}, which use the notion of auto-encoders to explicitly separate the feature representations private to each source/target domain from those that are shared between the domains. 
Although extending \textbf{DSN} to multiple domains might seem trivial, \textbf{DSN} requires an autoencoder per domain, making the model impractical in the case of more than a couple of domains. 
%In Sec.~\ref{conn}, we further discuss the differences between DSN and our model. 

The overall loss of \textbf{DSN} consists of a reconstruction loss for each domain modeled by a shared decoder, a similarity loss such as \textbf{MMD}, which encourages domain invariance modeled by a shared encoder, and a dissimilarity loss modeled by two private encoders: one for the source domain and one for the target domain.  While one could attempt to generalize \textbf{DSN} to multiple target domains by having individual per-target domain private encoders%, as illustrated in Fig.~\ref{model:pm}(right)
, doing so would prove problematic when the number of target domains is large --- each private encoder would require a large "private" dataset to learn the private parameters. Precisely, for multiple $(M)$ target domains, we could train a \textbf{DSN} model with one shared encoder, $M+1$ private encoder (one for each domain), and one shared decoder. This leads to $M+3$ models to train that implies the number of models increases linearly with the number of domains, as does the required training time.
Second, \textbf{DSN} uses an orthogonality constraint among the shared and the private representations which may not be strong enough to remove redundancy and enforce disentangling among different private spaces. Precisely, \textbf{DSN} defines the loss via a soft subspace orthogonality constraint between the private and shared representation of each domain. However, it does not enforce the private representation of different domains to be different that may result in redundancy of different private spaces. 

In addition, \textbf{DSN} enforces separation of spaces using the notion of Euclidean orthogonality,  e.g., $\|z_s - z_p\|^2$. In case of multiple target domains, this would result in learning of all pairs of private spaces independently. 
To address those deficiencies, we first explicitly couple different private encoders into a single private encoder model, $E_{\theta_p}$ of Fig.~\ref{model:pm}
%(left)
, which allows us to generalize to an arbitrary number of target domains. To assure that the information among the private and shared spaces is not shared (i.e., "orthogonal"), we define an information-theoretic criteria enforced by a domain classifier,$D_{\psi}$ of Fig.~\ref{model:pm}%(left)
, which aims to segment the private space into clusters that correspond to individual target domains. By using $D_{\psi}$ within the adversarial framework, \textbf{MTDA-ITA} learns simultaneously the shared and private features from different domains (see Fig.~\ref{vis}). We also show in Sec.~\ref{er} that our model performs better than the trivial extension of DSNs to the multi-domain case.
\section{Experimental Results}\label{er}
\begin{figure*}[t]
 	\begin{center}
    \subfloat[Digit datasets]{{\includegraphics[width=3.5cm, height=3.5cm]{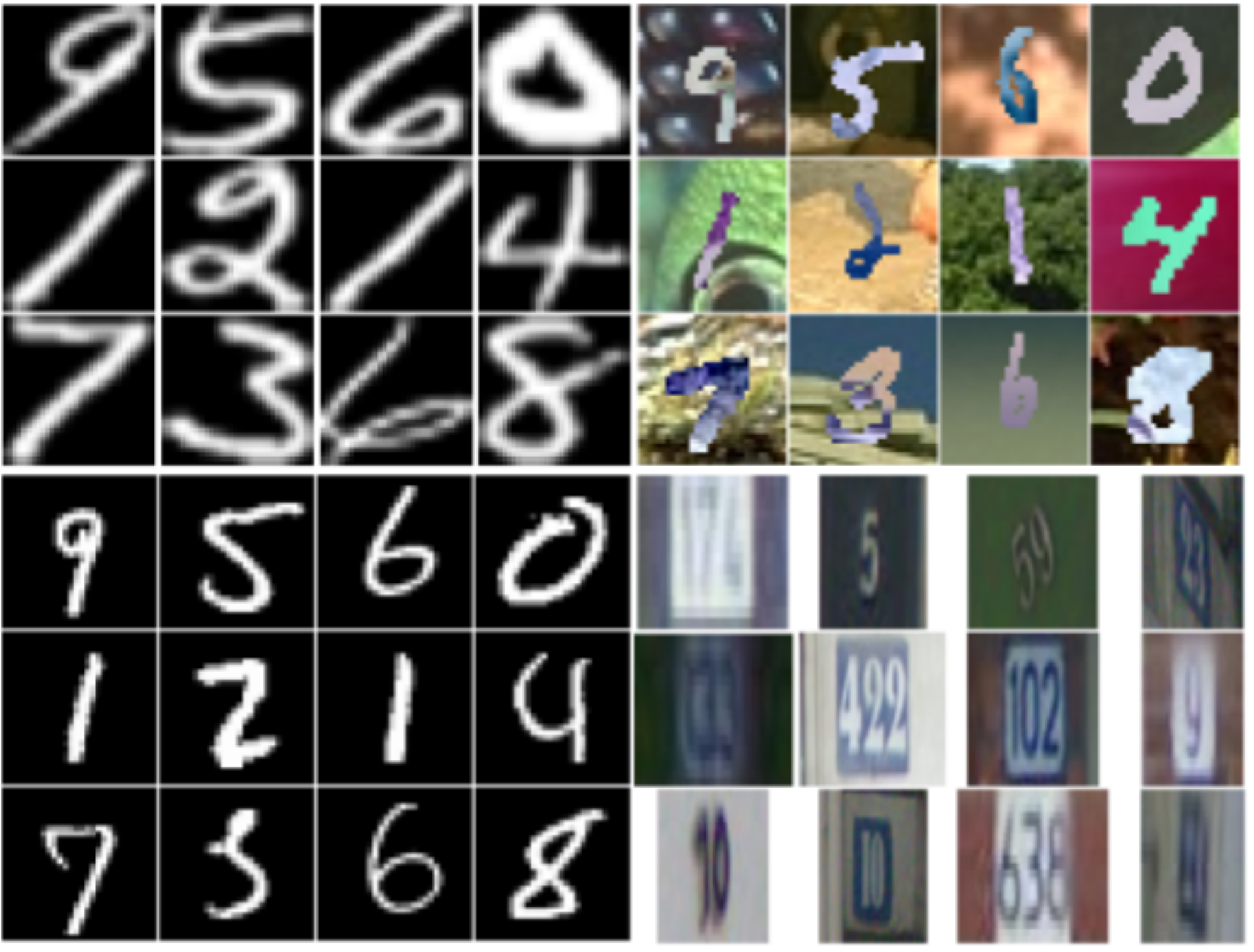} }}%
    \qquad
    \subfloat[PACS dataset]{{\includegraphics[width=3.5cm,height=3.5cm]{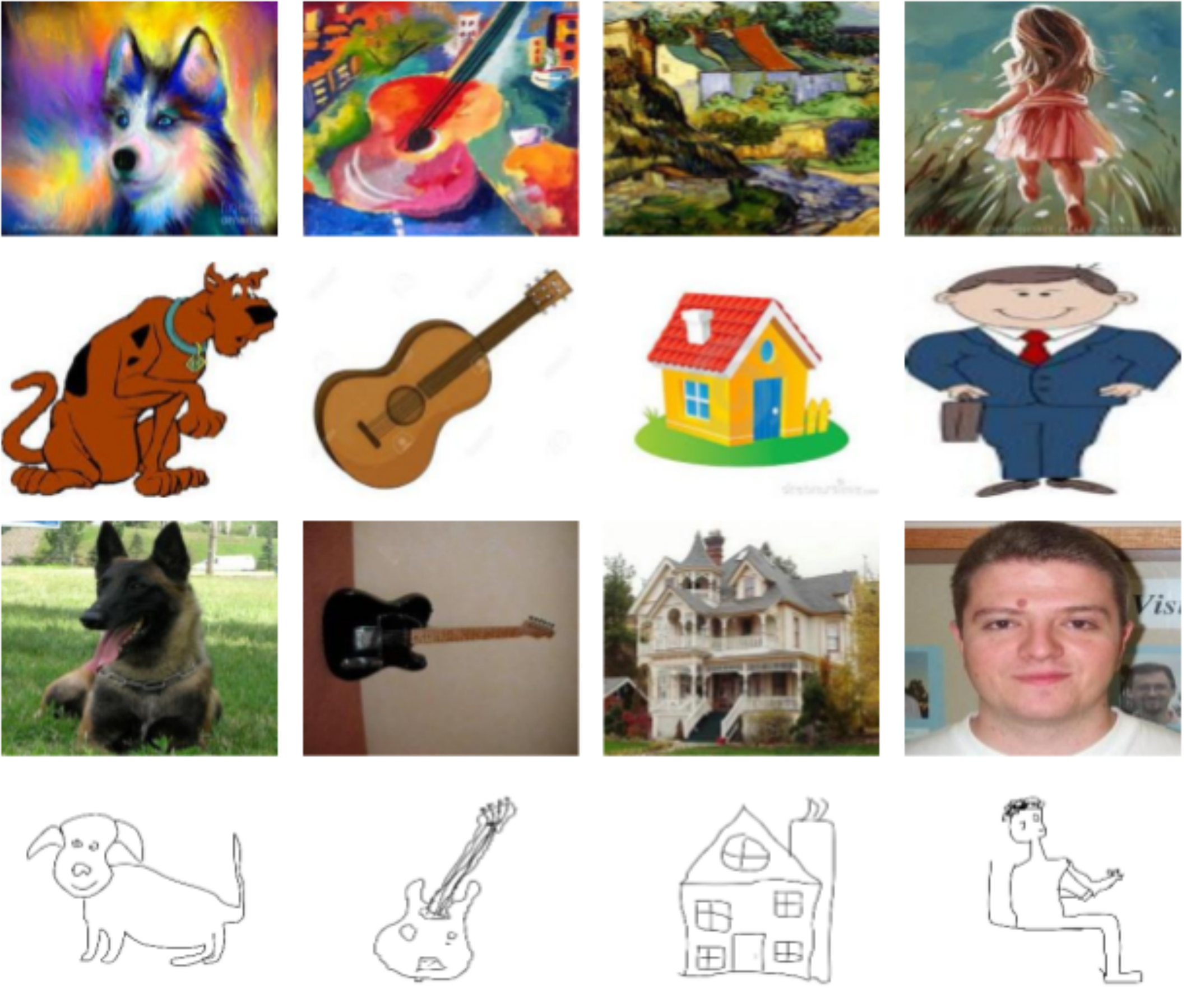} }}%
    \qquad
    \subfloat[Multi-PIE dataset]{{\includegraphics[width=3.5cm,height=3.5cm]{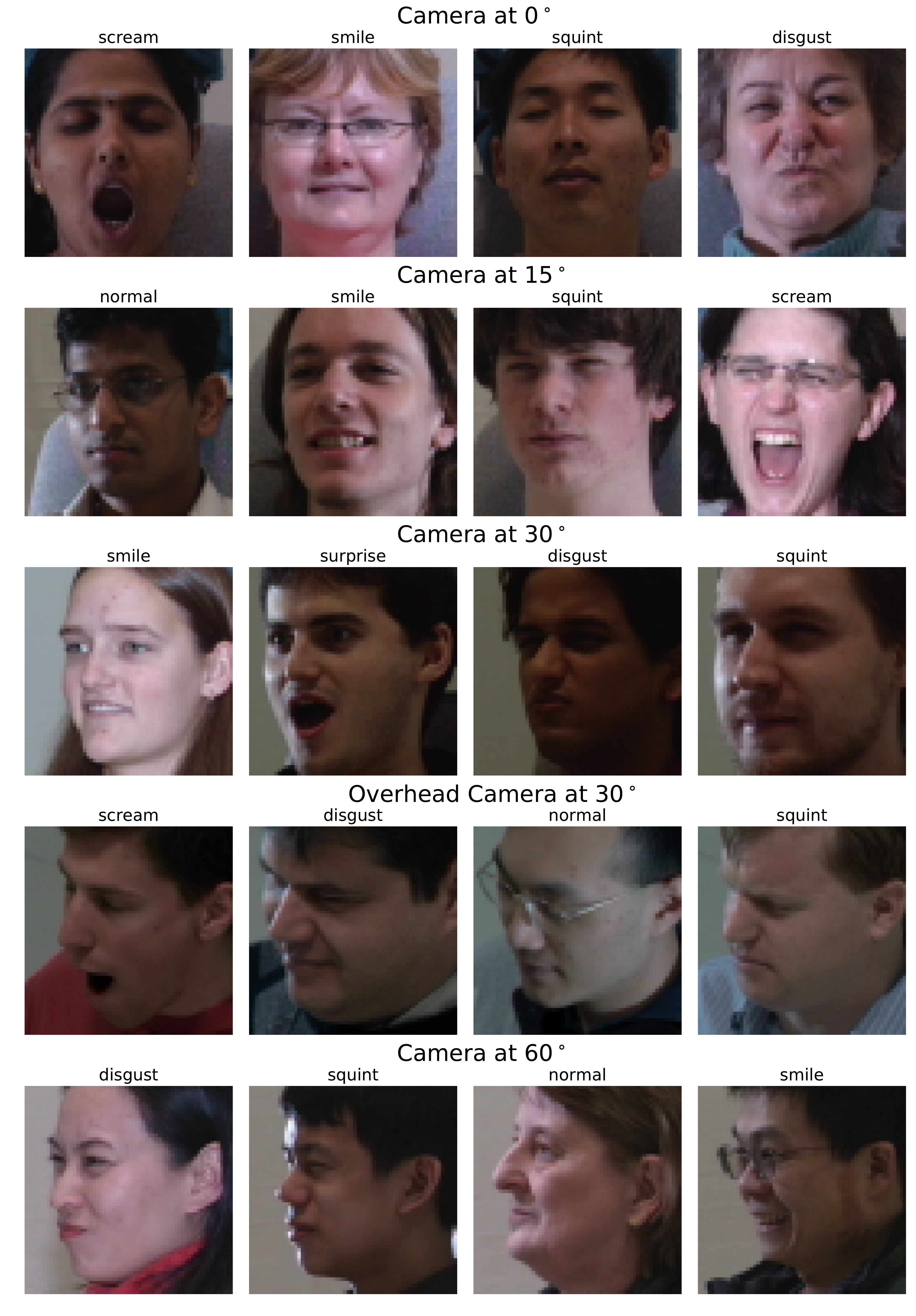} }}%{multipie-subject.pdf} }}%
    \label{fig:example}%
%  		\includegraphics[height=.18\textheight,width =.75\textwidth]{graphical_model.pdf}
 	\end{center}
 	%\caption{Exemplary images from different datasets. a) Digits datasets, b) Multi-PIE dataset (column 1: C11, column 2: C09, column 3: C05, column 4: C20, column 5: C24).}
    \caption{Exemplary images from different datasets. a) Digits datasets, b) PACS datatset (first row: Art-painting, second row: Cartoon, Third row: Photo, last row: Sketch), c) Multi-PIE dataset (each row corresponds to a different camera angle and each subject depicts an expression("normal", "smile", "surprise", "squint", "disgust", "scream") at every camera position).}
 	\label{amazon}
 \end{figure*}

We compare the proposed method with state-of-the-art methods on standard benchmark datasets: a
digit classification task that includes 4 datasets: \textbf{MNIST}~\cite{lecun1998gradient}, \textbf{MNIST-M}~\cite{ganin2016domain}, \textbf{SVHN}~\cite{netzer2011reading}, \textbf{USPS}~\cite{tzeng2017adversarial}, \textbf{Multi-PIE} expression recognition  dataset\footnote{\href{http://www.cs.cmu.edu/afs/cs/project/PIE/MultiPie/Multi-Pie/Home.html}{http://www.cs.cmu.edu/afs/cs/project/PIE/MultiPie/Multi-Pie/Home.html}}, and \textbf{PACS} multi-domain image recognition benchmark~\cite{li2017deeper},
a new dataset designed for the cross-domain recognition
problems. Figure~\ref{amazon} illustrates image samples from different datasets and domains. We evaluate the performance of all methods
with classification accuracy metric. 

We used ADAM~\cite{kingma2014adam} for training; the learning rate was set to $0.0002$ and momentums to $0.5$ and $0.999$. We used batches of size $16$ from each domain, and the input
images were mean-centered/rescaled to $[-1,1]$. The hyper-parameters are empirically set as $\lambda_r =1.0, \lambda_c = 0.01, \lambda_d = 0.20$. 

For the network architecture,
our private/shared encoders consisted of three convolutional layers as the front-end and four basic residual blocks as
the back-end. The decoder consisted of four basic residual blocks as the front-end and three transposed convolutional layers as the back-end. The discriminator and the classifier consisted of stacks of convolutional layers. We used ReLU for nonlinearity. TanH function is
used as the activation function of the last layer in the decoder $F$ 
for scaling the output pixels to $[-1,1]$. The details of the networks are given in Appendix. 

The  quantitative  evaluation  involves  a  comparison
of the performance of our model to previous work and to
“\textbf{Source Only}” and “\textbf{1-NN}” baselines that do not use
any  domain  adaptation. For "\textbf{Source Only}" baseline, we  train  models  only  on  the  unaltered  source  training  data  and  evaluate on the target test data.
We compare the proposed method \textbf{MTDA-ITA} with several related methods designed for pair-wise source-target adaptation: \textbf{CORAL}~\cite{sun2016deep}, \textbf{DANN}~\cite{ganin2014unsupervised}, \textbf{ADDA}~\cite{tzeng2017adversarial}, \textbf{DTN}~\cite{zhang2015deep}, \textbf{UNIT}~\cite{liu2017unsupervised}, \textbf{PixelDA}~\cite{bousmalis2017unsupervised}, and \textbf{DSN}~\cite{bousmalis2016domain}. We reported the results of two following baselines: (i) one is to combine all the target domains
into a single one and train it using \textbf{MTDA-ITA}, which we denote as (\textbf{c-MTDA-ITA}). (ii) the other one is to
train multiple \textbf{MTDA-ITA} separately, where each one corresponds to a source-target pair which we denote as (\textbf{s-MTDA-ITA}). For completeness, we reported the results of the competing methods by combining all the target domains
into a single one (denoted by \textbf{c-DTN}, \textbf{c-ADDA}, and \textbf{c-DSN}) as well. We also extend \textbf{DSN} to multiple domains by adding multiple private encoder to it (denoted by \textbf{mp-DSN}) and contrast it with our model. %The methods are also compared with 
%the no Adaptation baseline, which is 
%the results obtained by the Nearest Neighbor (\textbf{1-NN}) classifier on the source domain data without adaptation. 

\subsection{Digits Datasets}
We combine four popular digits datasets (\textbf{MNIST},
\textbf{MNIST-M}, \textbf{SVHN}, and \textbf{USPS}) to build the multi-target domain dataset.  All images were uniformly rescaled to $32\times 32$. We take each of \textbf{MNIST-M}, \textbf{SVHN}, \textbf{USPS}, and \textbf{MNIST} as source domain in turn, and the rest as targets. We use all labeled source images and all  unlabeled  target  images, following  the  standard  evaluation  protocol  for  unsupervised  domain  adaptation~\cite{ganin2016domain,long2016deep}. 

We show the accuracy of different methods in Tab.~\ref{1}. The results show that first of all \textbf{cMTDA-ITA} has worse performance than \textbf{sMTDA-ITA} and \textbf{MTDA-ITA}. We have similar
observations for \textbf{ADDA}, \textbf{DTN}, and \textbf{DSN} that demonstrates a naive combination of different target datasets can sometimes even decrease the performance of the competing methods. Moreover, \textbf{MTDA-ITA} outperforms the state-of-the-art methods in most of domain transformations. The higher performance of \textbf{MTDA-ITA} compared to other methods is mainly attributed to the joint adaptation of related domains where each domain could benefit of other related domains. Furthermore, from the results obtained, we see that it is beneficial to use information coming from unlabeled target data (see Eq.~\ref{theta-c} for updating the classifier $C_{\theta_c}$) during the learning process, compared to when no data from target domain is used (See the ablation study section for more information). Indeed, using our scheme, we find a representation space in which embeds the knowledge from the target domain into the learned classifier. By contrast, the competing methods do not provide a principled way of sharing information across all domains, leading to overall lower performance. The results also verify the superiority of \textbf{MTDA-ITA} over \textbf{mp-DSN}. This can be due to (i) having multiple private encoders increase the number of parameters that may lead to model overfitting, (ii) superiority of the \textbf{MTDA-ITA}'s domain adversarial loss over the \textbf{DSN}'s MMD loss to separate the shared and private features, (iii) utilization of the unlabeled target data to regularize the classifier in \textbf{MTDA-ITA}.          
\begin{table*}[t]
	\caption{Mean classification accuracy on digit classification. M: MNIST; MM: MNIST-M, S: SVHN, U: USPS. The best is shown in red. c-X: combining all target domains into a single one and train it using X. s-MTDA-ITA: training multiple MTDA-ITA where each one correspond to a source-target pair. mp-DSN: extended DSN with multiple private encoder. *UNIT trains with the extended SVHN ($>500$K images vs ours $72$K). *PixelDA uses ($\approx 1,000$) of labeled target domain data as a validation set for tuning the hyperparameters.}
	\centering
	\resizebox{1\columnwidth}{!}{%
		\begin{tabular}[b]{c||c|c|c|c|c|c|c|c|c|c|c|c}
			\hline
			method & S $\rightarrow$ M & S $\rightarrow$ MM& S $\rightarrow$ U  & M $\rightarrow$ S & M $\rightarrow$ MM& M $\rightarrow$ U & MM $\rightarrow$ S& MM $\rightarrow$  M&MM $\rightarrow$ U &U $\rightarrow$ S &U $\rightarrow$ M &U $\rightarrow$  MM \\
			\hline
			\textbf{Source Only} &62.10 &40.43 &39.90 &30.29 &55.98 &78.30 &40.00 &84.46 &80.43  &23.41 &50.64 & 41.45\\
			\textbf{1-NN} & 35.86& 18.21&29.31 & 28.01& 12.58& 41.22& 21.45& 82.13& 36.90 &15.34 &38.45 &18.54 \\
			%\hline
			\textbf{CORAL}~\cite{sun2016deep}  & 63.10& 54.37& 50.15 &33.40 &57.70 &81.05  & 40.20& 84.90& 87.54&38.90 &85.01 &60.45\\
			\textbf{DANN}\cite{ganin2016domain} &73.80 &61.05 &62.54 &35.50 &77.40 &81.60 &51.80 &61.05 &85.34 &35.50 &77.40 &61.60  \\
			\hline
			\textbf{ADDA}\cite{tzeng2017adversarial} & 77.68& 64.23&64.10  &30.04 &91.47 &90.51& 40.64& 92.82&80.70  &41.23 &90.10 &56.21  \\
			\textbf{c-ADDA} &80.10 & 56.80& 64.80 & 27.50& 83.30& 84.10&35.43 &88.47 &74.19  &39.36 &84.67 &52.54  \\
			\hline
			\textbf{DTN}\cite{zhang2015deep} &81.40 &63.70 &60.12 &40.40 &85.70 &85.80 &48.80 &88.80 &90.68 &42.43 &89.04 &55.78\\
			\textbf{c-DTN} &82.10 &59.30 &56.87  &38.32 &80.90 & 79.31&44.21 &83.60 &84.98  &39.75 &85.04 &48.86  \\
			\hline
			\textbf{PixelDA}\cite{bousmalis2017unsupervised} &-- &-- &-- &-- &\textbf{\color{red}{98.10}}$^*$ &\textbf{\color{red}{94.10}}$^*$ &-- &-- &-- &-- &-- &-- \\
            \textbf{UNIT}\cite{liu2017unsupervised} &\textbf{\color{red}{90.6}}$^*$ &-- &-- &-- & --&92.90 &-- &-- &-- & &90.60 &-- \\
            \hline
			\textbf{DSN}\cite{bousmalis2016domain} & 82.70&64.80 &65.30 &49.30 &83.20 &91.65 & 51.50&90.20 &89.95 &\textbf{\color{red}{48.20}} &\textbf{\color{red}{91.40}} &60.45 \\
			\textbf{c-DSN} & 83.10& 60.56 & 60.35 &46.80 &80.49 & 88.21& 47.10&84.60 &84.80  &40.50 &86.05 &56.25  \\
			\textbf{mp-DSN} &83.40 & 61.00 & 58.10 & 47.35& 79.30 &78.45 & 47.15 & 85.51 &83.24  &38.30 &87.40 & 55.47 \\
			\hline
            \textbf{s-MTDA-ITA} & 82.90 & 63.10 & 63.54 & 49.60 & 82.42 & 89.21 & 50.55 & 94.82 & 89.05 & 40.13 & 87.10 & 61.01\\
            \textbf{c-MTDA-ITA} & 79.20 & 59.90 & 63.70 & 45.30 & 77.12 & 87.47 & 47.32 & 90.20 & 90.01 & 41.10 & 85.35 & 60.31\\
			\textbf{MTDA-ITA} & \textbf{\color{red}{84.60}}&\textbf{\color{red}{65.30}} &\textbf{\color{red}{70.03}}  &\textbf{\color{red}{52.01}} &85.50 &\textbf{\color{red}{94.20}} &\textbf{\color{red}{53.50}}&\textbf{\color{red}{98.20}} &\textbf{\color{red}{94.10}} &46.00 &\textbf{\color{red}{91.50}} &\textbf{\color{red}{67.30}}  			%	\hline   
		\end{tabular}
	}
	\label{1}
\end{table*}
\begin{table*}[t]
	\caption{Mean classification accuracy on Multi-PIE classification. The best is shown in red.}
	\centering
	\resizebox{1\columnwidth}{!}{%
		\begin{tabular}[b]{c||c|c|c|c|c|c|c|c|c|c|c|c}
			\hline
			method & C05 $\rightarrow$ C08 & C05 $\rightarrow$ C09& C05 $\rightarrow$ C13  & C05 $\rightarrow$ C14 & C13 $\rightarrow$ C05& C13 $\rightarrow$ C08 & C13 $\rightarrow$ C09& C13 $\rightarrow$  C14&C14 $\rightarrow$ C05 &C14 $\rightarrow$ C08 &C14 $\rightarrow$ C09 &C14 $\rightarrow$  C13 \\
			\hline
			\textbf{Source Only} &31.56 &40.67 &39.89 &54.70 &50.79 &45.90 &40.04 &59.68 &60.03& 36.80 &40.11&60.57 \\
			\textbf{1-NN} &27.28 &31.22 &33.66 &47.04 &33.21 &37.01 &34.45 &48.79 &47.44 &28.24 &30.86 &44.86 \\
			%\hline
			\textbf{CORAL}~\cite{sun2016deep}  &36.55&38.60 &40.60 &55.29 &54.89 &48.90 &40.30 &68.90&59.98 &40.63 &40.80 &65.11 \\
			\textbf{DANN}\cite{ganin2016domain} &40.30 &41.20 &40.12 & 58.90&57.86 &50.30 &45.30 &70.68&57.20 &40.22 &40.77 &70.50   \\
			\hline
			\textbf{ADDA}\cite{tzeng2017adversarial} &33.21 &30.8 6&52.44 &70.18 &64.83 &63.20 &55.48 &74.25 &73.62 &43.56 &38.68 &72.84 \\
			\textbf{c-ADDA} &46.88 & 36.38 & 39.14  & 65.41 & 59.20& 30.70&53.20 &68.33 & 65.88 &30.60 &45.34 &64.30  \\
			\hline
			\textbf{DTN}\cite{zhang2015deep} &38.50 &30.56 &55.78 &68.90 &63.78 &60.45 &60.55 &72.60&70.67 &41.55 &41.45 &70.67 \\
			\textbf{c-DTN} &41.70 &31.10 &50.19  &60.34 & 57.53&55.24 &57.14 &65.16 & 63.80 & 38.97& 39.80&62.10  \\
			\hline
            \textbf{PixelDA}\cite{bousmalis2017unsupervised} &44.93 &44.75 &45.18 &46.88 &45.68 &44.95 &44.45 &\textbf{\color{red}{90.50}} &46.28 &45.89 &44.45 &69.15 \\
            \textbf{UNIT}\cite{liu2017unsupervised} &44.47 &44.47 &44.47 &44.51 &44.14 &44.47 &44.21 &44.47 &43.03 &44.44 &44.47 &44.47 \\
            \hline
			\textbf{DSN}\cite{bousmalis2016domain} &45.12 &44.35 &48.12 &75.00 &64.15 &57.70 &49.15 &80.75 &82.20 &38.75 &45.00 &80.50 \\
			\textbf{c-DSN} & 42.52&38.54 &34.15  &69.45 &57.34 &31.63 &51.17 &74.52 &82.01  &34.25 &42.63 &79.42  \\
			\textbf{mp-DSN}\cite{tzeng2017adversarial} &41.30 &35.14  &34.40  &65.70 &55.20 &30.40 &47.80 &75.30 &80.75  &30.20 &43.00 &79.02  \\
			\hline
            \textbf{s-MTDA-ITA} &44.40 &44.60 &47.65 &80.20 &70.10 &58.90 &58.10 &80.12&82.05 &45.90 &52.67 &81.60 \\
            \textbf{c-MTDA-ITA} &40.49 &40.70 &42.80 &71.60 &60.34 &55.67 &57.10 &73.50&76.80 &43.10 &48.10 &80.90 \\
			\textbf{MTDA-ITA} &\textbf{\color{red}{49.01}} & \textbf{\color{red}{48.23}} &\textbf{\color{red}{53.13}} &\textbf{\color{red}{84.29}} &\textbf{\color{red}{78.40}} & \textbf{\color{red}{66.70}}&\textbf{\color{red}{70.30}} &85.49&\textbf{\color{red}{87.20}} & \textbf{\color{red}{61.40}}& \textbf{\color{red}{60.05}} &\textbf{\color{red}{86.70}} \\
           % & \textbf{\color{red}{84.60}}&\textbf{\color{red}{65.30}} &\textbf{\color{red}{70.03}}  &\textbf{\color{red}{52.01}} &85.50 &\textbf{\color{red}{94.20}} &\textbf{\color{red}{53.50}}&\textbf{\color{red}{98.20}} &\textbf{\color{red}{94.10}} &46.00 &\textbf{\color{red}{91.50}} &\textbf{\color{red}{67.30}}  			
		\end{tabular}
	}
	\label{pie}
\end{table*}
\begin{table*}[t]
	\caption{Mean classification accuracy on Multi-PIE classification. The best (red).}
	\centering
	\resizebox{1\columnwidth}{!}{%
		\begin{tabular}[b]{c||c|c|c|c|c|c|c|c}
			\hline
			method & C08 $\rightarrow$ C05 & C08 $\rightarrow$ C09& C08 $\rightarrow$ C13  & C08 $\rightarrow$ C14 & C09 $\rightarrow$ C05& C09 $\rightarrow$ C08 & C09 $\rightarrow$ C13& C09 $\rightarrow$  C14 \\
			\hline
			\textbf{Source Only} & 33.70& 50.10& 50.80& 40.13& 33.32& 48.24&49.24 &36.19 \\
			\textbf{1-NN} &28.75 &35.39 &39.79 &32.13 &26.82 &35.30 &34.26 &28.41 \\
			%\hline
			\textbf{CORAL}~\cite{sun2016deep}  &35.89 &55.79 &60.00 &40.67 &35.89 &51.56 &50.45 &40.67 \\
			\textbf{DANN}\cite{ganin2016domain} &40.20 &56.89 &55.83 &43.25 &50.63 &\textbf{\color{red}{58.40}} &55.81 &48.90 \\
			\hline
			\textbf{ADDA}\cite{tzeng2017adversarial} &37.40 &58.40 &60.40 &42.10 &29.40 &53.30 &45.30 &38.30\\
			\textbf{c-ADDA} &41.60 &39.65 &50.00  &46.25 &45.01 &52.14 &37.43 &43.26 \\
			\hline
			\textbf{DTN}\cite{zhang2015deep} &44.13 &57.42 &55.89 &45.76 &44.53 &57.34 &52.43 &51.55 \\
			\textbf{c-DTN} &45.10 &49.78 &47.43  &45.79 &49.80 &55.69 &50.10 &52.31  \\
			\hline
            \textbf{PixelDA}\cite{bousmalis2017unsupervised} &\textbf{\color{red}{46.45}} &44.33 &44.87 &46.83 &45.63 &16.37 &45.43 &47.00 \\
			\hline
            \textbf{UNIT}\cite{liu2017unsupervised} &43.88 &43.99 &44.47 &44.47 &44.47 &43.95 &44.64 &44.47 \\
            \hline
			\textbf{DSN}\cite{bousmalis2016domain} &\textbf{\color{red}{46.25}} &47.50 &\textbf{\color{red}{62.15}} &39.72 &45.85 &56.65 &56.5 &42.87 \\
			\textbf{c-DSN} & 45.82& 44.64& 45.60 &46.32 &45.18 &45.52 &44.79 &47.37  \\
			\textbf{mp-DSN} &42.19 &44.70  &42.47  &40.50 &45.00 &43.80 &45.79 &42.39   \\
			\hline
            \textbf{s-MTDA-ITA} &44.77 &45.61 &60.00 &46.70 &49.06 &55.33 &59.90 &50.64 \\
            \textbf{c-MTDA-ITA} &44.35 &42.67 &58.90 &44.32 &46.74 &54.11 &56.89 &49.64 \\
			\textbf{MTDA-ITA} &\textbf{\color{red}{46.30}} &\textbf{\color{red}{60.60}} &60.50 &\textbf{\color{red}{50.40}} &\textbf{\color{red}{55.59}} &57.80 &\textbf{\color{red}{64.20}} &\textbf{\color{red}{56.34}} \\
            %& \textbf{\color{red}{84.60}}&\textbf{\color{red}{65.30}} &\textbf{\color{red}{70.03}}  &\textbf{\color{red}{52.01}} &85.50 &\textbf{\color{red}{94.20}} &\textbf{\color{red}{53.50}}&\textbf{\color{red}{98.20}}  			%	\hline   
		\end{tabular}
	}
	\label{pie2}
\end{table*}
\begin{table*}[t]
	\caption{Mean classification accuracy on PACS dataset classification. A:Art-painting, C:Cartoon, S:Sketch, P:Photo. The best (red).}
	\centering
	\resizebox{0.7\columnwidth}{!}{%
		\begin{tabular}[b]{c||c|c|c|c|c|c}
			\hline
			method & P $\rightarrow$ A & P $\rightarrow$ C& P $\rightarrow$ S   & A $\rightarrow$ P & A$\rightarrow$ C& A $\rightarrow$ S  \\
			\hline
			\textbf{1-NN} & 15.28& 18.16& 25.60&22.70 &19.75 & 22.70 \\
			%\hline
			\textbf{ADDA}\cite{tzeng2017adversarial} & 24.35&20.12&22.45 & 32.57&17.68 & 18.90\\
			\textbf{DSN}\cite{bousmalis2016domain} &28.42 &21.14 &25.64 &29.54 & 25.89 & 24.69\\
			\hline   
            \textbf{s-MTDA-ITA} &28.02 &21.64 &26.24 &31.06 & 25.09 & 25.89\\
            \textbf{c-MTDA-ITA} &25.35 &20.24 &23.64 &26.54 & 20.30& 22.38\\
		\textbf{MTDA-ITA} &\textbf{\color{red}{31.40}} &\textbf{\color{red}{23.05}} &\textbf{\color{red}{28.24}} &\textbf{\color{red}{35.74}} & \textbf{\color{red}{27.00}} & \textbf{\color{red}{28.90}}\\

		\end{tabular}%
	}
	\label{pacs}
\end{table*}

\subsection{\textbf{Multi-PIE} dataset}
The \textbf{Multi-PIE} dataset includes face images of $337$ individuals captured from different expressions, views, and illumination conditions (Fig.~\ref{amazon}(c)). 
For this experiment, we use $5$ different camera views (positions) $C05$, $C08$, $C09$, $C13$, and $C14$ as different domains (Fig.~\ref{amazon}(c)) and the face expressions (normal, smile, surprise, squint, disgust, scream) as labels. Each domain contains $27120$ images of size $64\times 64\times  3$.
We used each view as the source domain, in turn, and the rest as targets. We expect the face inclination angle to reflect the complexity of transfer learning. 
 Tables~\ref{pie} and~\ref{pie2} shows the classification accuracy for $C05,C08,C09,C13$, and $C14$ as source domain. As can be seen, \textbf{MTDA-ITA} achieves the best performances as well as the best scores in most settings that verifies the effectiveness of \textbf{MTDA-ITA} for multi-target domain adaptation. Clearly, with the increasing camera angle, the image structure changes up to a certain extent (the views become heterogeneous). However, our method produces better results even under such very challenging conditions.
\subsection{\textbf{PACS} dataset}
This dataset contains $9991$ images $(227 \times 227 \times 3\; \text{dimension})$
across $7$ categories (‘dog’, ‘elephant’, ‘giraffe’,‘guitar’, ‘house’, ‘horse’ and ‘person’) and 4 domains of different
stylistic depictions (‘Photo’, ‘Art painting’, ‘Cartoon’
and ‘Sketch’). The very diverse depiction styles provide a significant
gap between domains, coupled with the small number of data samples, making it extremely challenging for domain adaptation.  Consequently, the dataset was originally used for multi-source to single target domain adaptation~\cite{li2017deeper}. 
Instead, we tackle a significantly more challenging problem of single-source to multiple target adaptation. Tab.~\ref{pacs} shows the classification accuracy of various methods. \textbf{MTDA-ITA} consistently achieves the best performance for all transfer tasks. Evaluations were obtained by training all models (\textbf{ADDA}, \textbf{DSN}, and ours) from scratch on the \textbf{PACS} dataset.  Note that the overall performance figures are low due to the extreme difficulty of the transfer task, induced by large differences among domains. 

\subsection{Ablation Studies}
We performed an ablation study on the proposed model measuring impact of various terms on the model's performance. To this end, we conducted additional experiments for the digit datasets with different components ablation, i.e., training
without the reconstruction loss (denoted as \textbf{MTDA-woR}) by setting $\lambda_r=0$, training without the classifier entropy loss (denoted as \textbf{MTDA-woE}) by setting $\lambda_c=0$, training without the multi-domain separation loss (denoted as \textbf{MTDA-woD}) by setting $\lambda_d=0$. 

\begin{figure*}[t]
 	\begin{center}
    \subfloat[Source domain:\textbf{SVHN}]{\label{fig:abl_svhn}{\includegraphics[width=5.7cm,height=3cm]{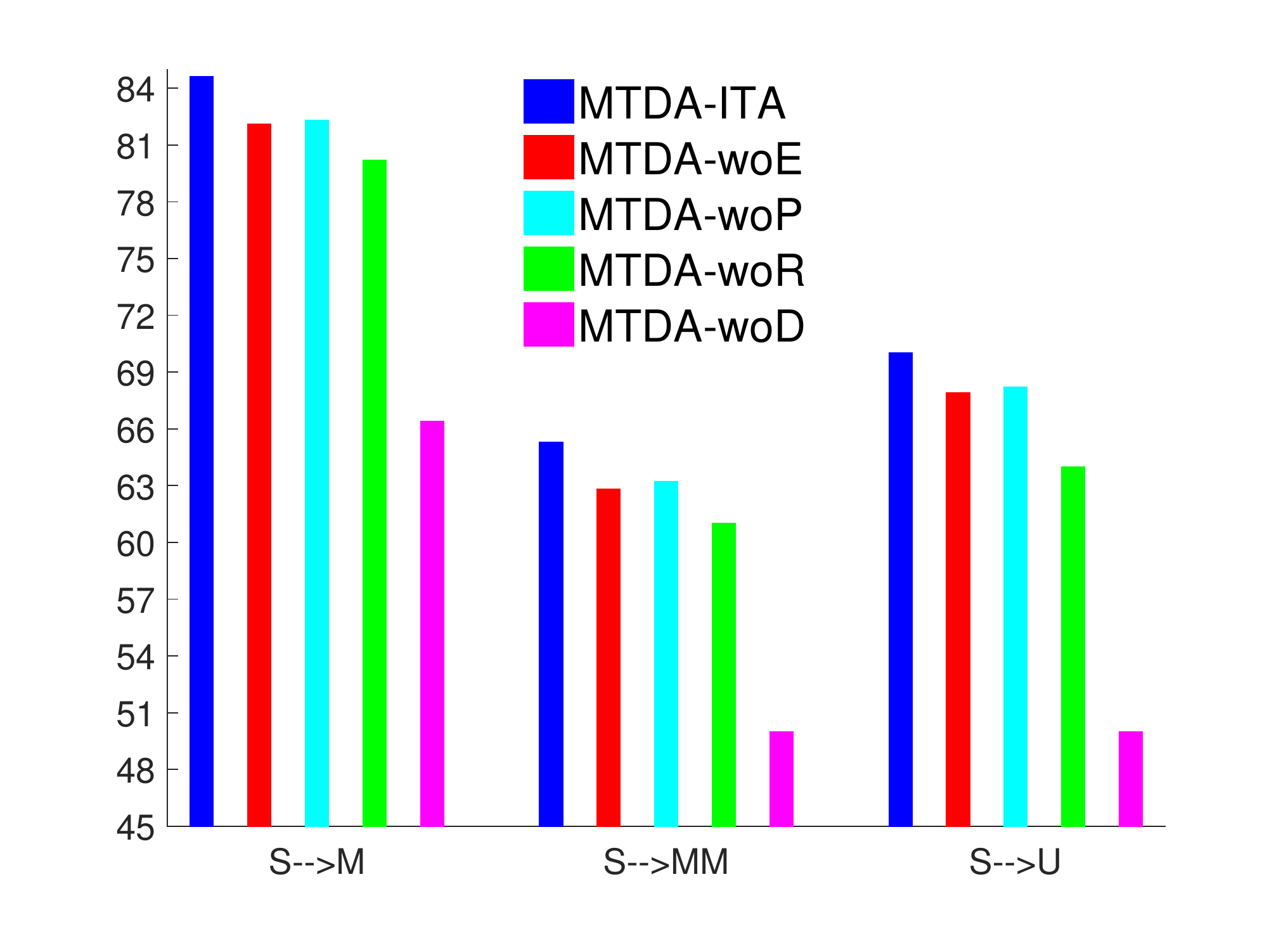} }}%
    \qquad
    \subfloat[Source domain:\textbf{MNIST-M}]{\label{fig:abl_mnistm}{\includegraphics[width=5.7cm,height=3cm]{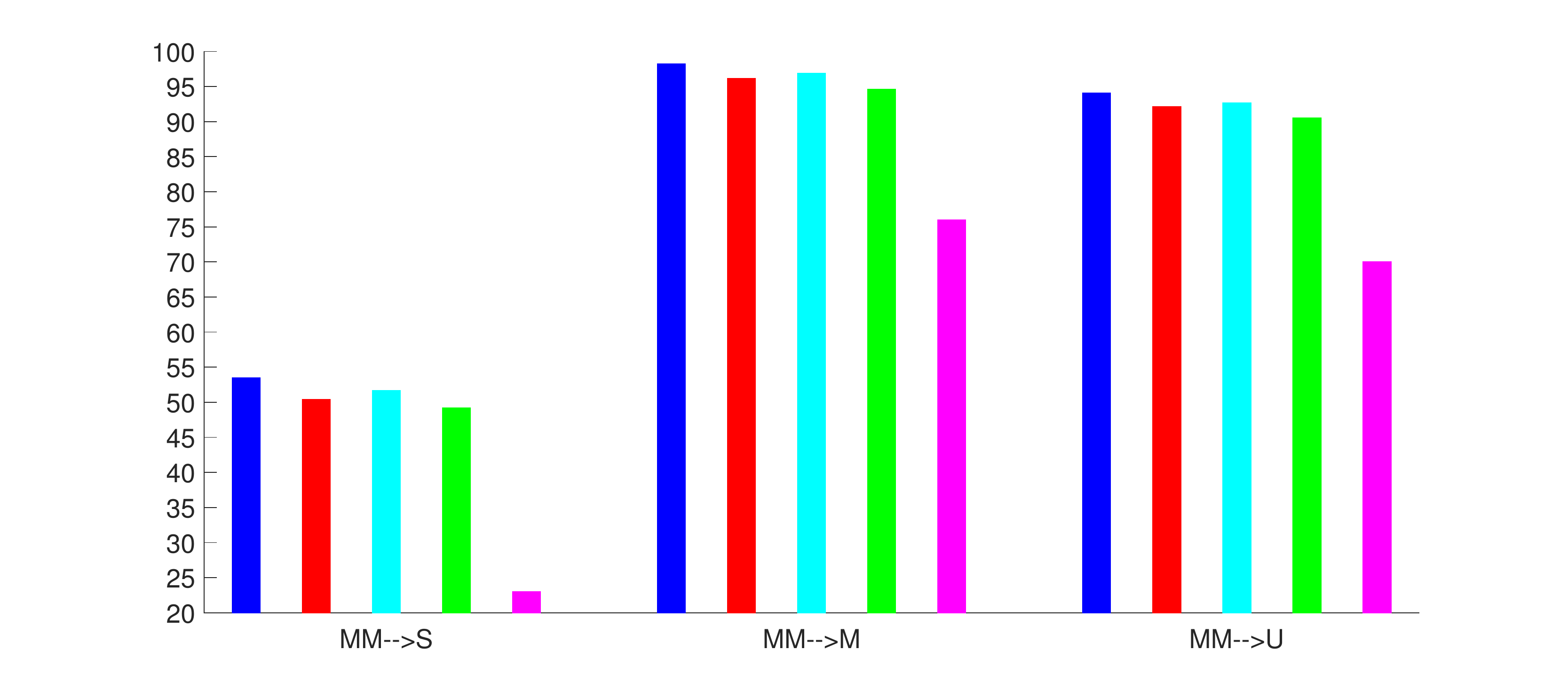} }}%
    
 	\end{center}
    \begin{center}
    \subfloat[Source domain: \textbf{MNIST}]{\label{fig:abl_mnist}{\includegraphics[width=5.7cm,height=3cm]{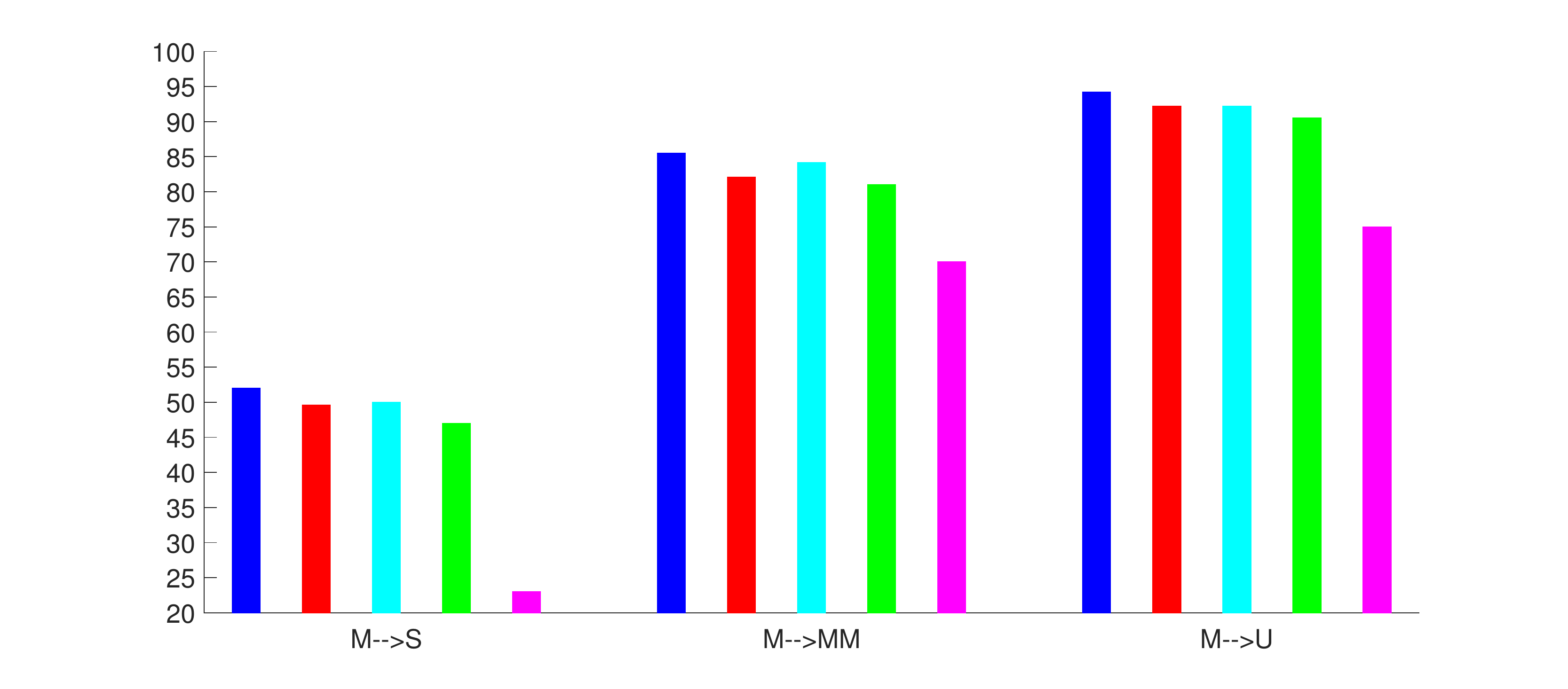} }}%
    \qquad
    \subfloat[Source domain: \textbf{USPS}]{\label{fig:abl_usps}{\includegraphics[width=5.7cm,height=3cm]{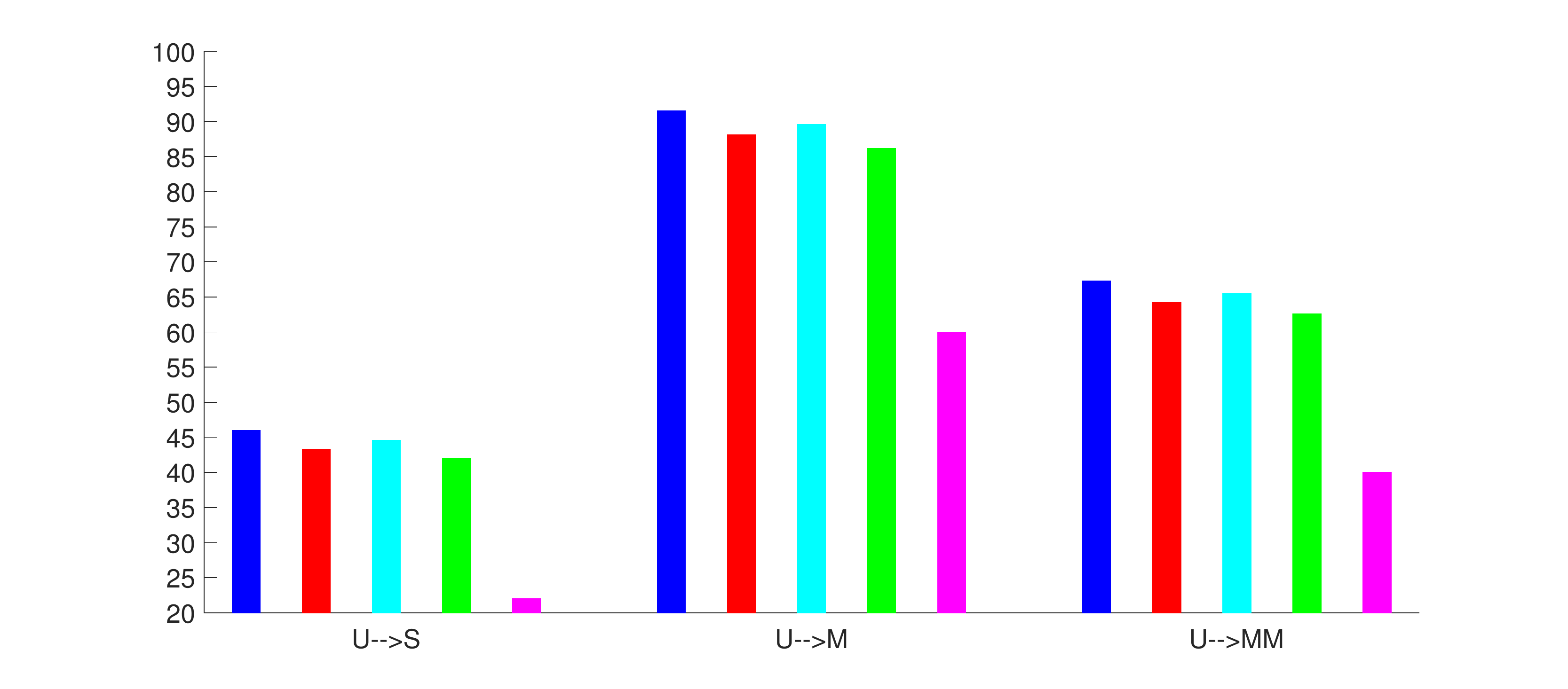} }}%
   
 	\end{center}
 	
    \caption{Ablation of \textbf{MTDA-ITA} on Digit dataset.
We show that each component of our method,
Reconstruction loss, Classifier entropy loss with separating shared/private features, contributes to the overall performance.}
 	\label{ablation}
 \end{figure*}

As can be seen from Fig.~\ref{ablation}, disabling each of the above components leads to degraded performance. More precisely, the average drop by disabling the classifier entropy loss is $\approx 3.5\%$. Similarly, by disabling the reconstruction loss and the multi-domain separation loss, we have $\approx 4.5\%$ and $\approx 22\%$ average drop in performance, respectively. Clearly, by disabling the multi-domain separation loss, the accuracy  drops significantly due to the severe data distribution mismatch between different domains. The figure also demonstrates that leveraging the unlabeled data from multiple target domains during training enhances the generalization ability of the model that leads to higher performance. In addition, the performance drop caused by removing the reconstruction loss , i.e., without the private encoder/decoder, indicates (i) the benefit of modeling the latent features as the combination of shared and private features, (ii) the ability of the model's domain adversarial loss to effectively learn those features. 

In order to examine the effect of the private features on the model's classification performance, we took the \textbf{MTDA-ITA} and trained it without the private encoder (denoted as \textbf{MTDA-woP}). As Fig.~\ref{ablation} shows, without the private features, the model performed consistently worse ($\approx 2\%$ average drop in performance) in all scenarios. This demonstrates explicitly modeling what is unique to each domain can improve the model’s ability to extract domain–invariant features. 

In summary, this ablation
study showed that the individual components bring complimentary information
to achieve the best classification results.
 \subsection{Feature Visualization}
\begin{figure}[H]
     \centering
     \begin{tabular}{ccc}
     & Domains & Classes \\
     \rotatebox[origin=c]{90}{No adaptation} & 
     \subfloat[]{\label{fig:embed_d_orig}\includegraphics[trim={3.2cm 2.5cm 0cm 0cm},clip,height = .16\linewidth,width=4.65cm,valign=c]{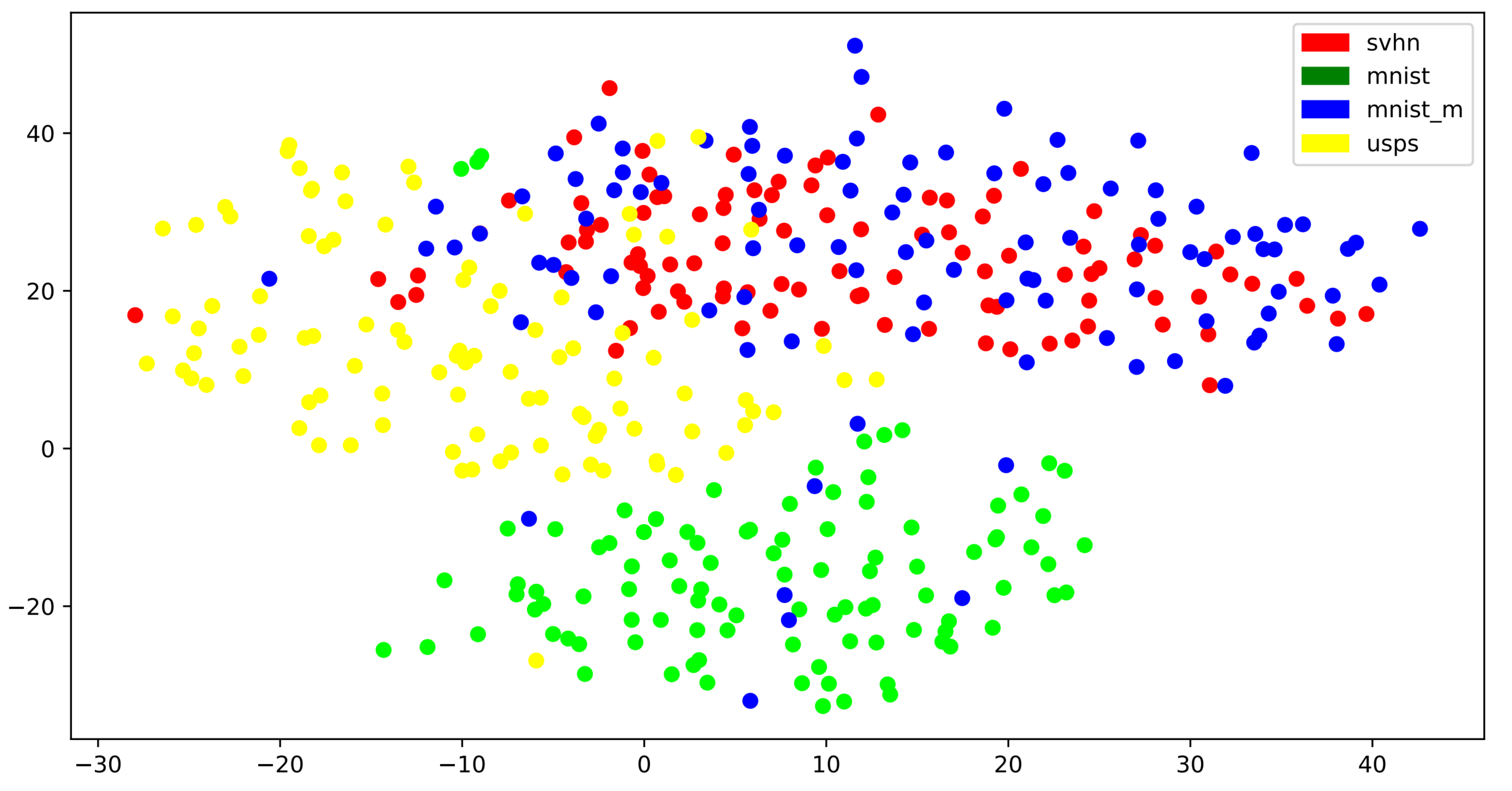}} &
	 \subfloat[]{\label{fig:embed_c_orig}\includegraphics[trim={3.2cm 2.3cm 0cm 0cm},clip,height = .16\linewidth,width=4.65cm,valign=c]{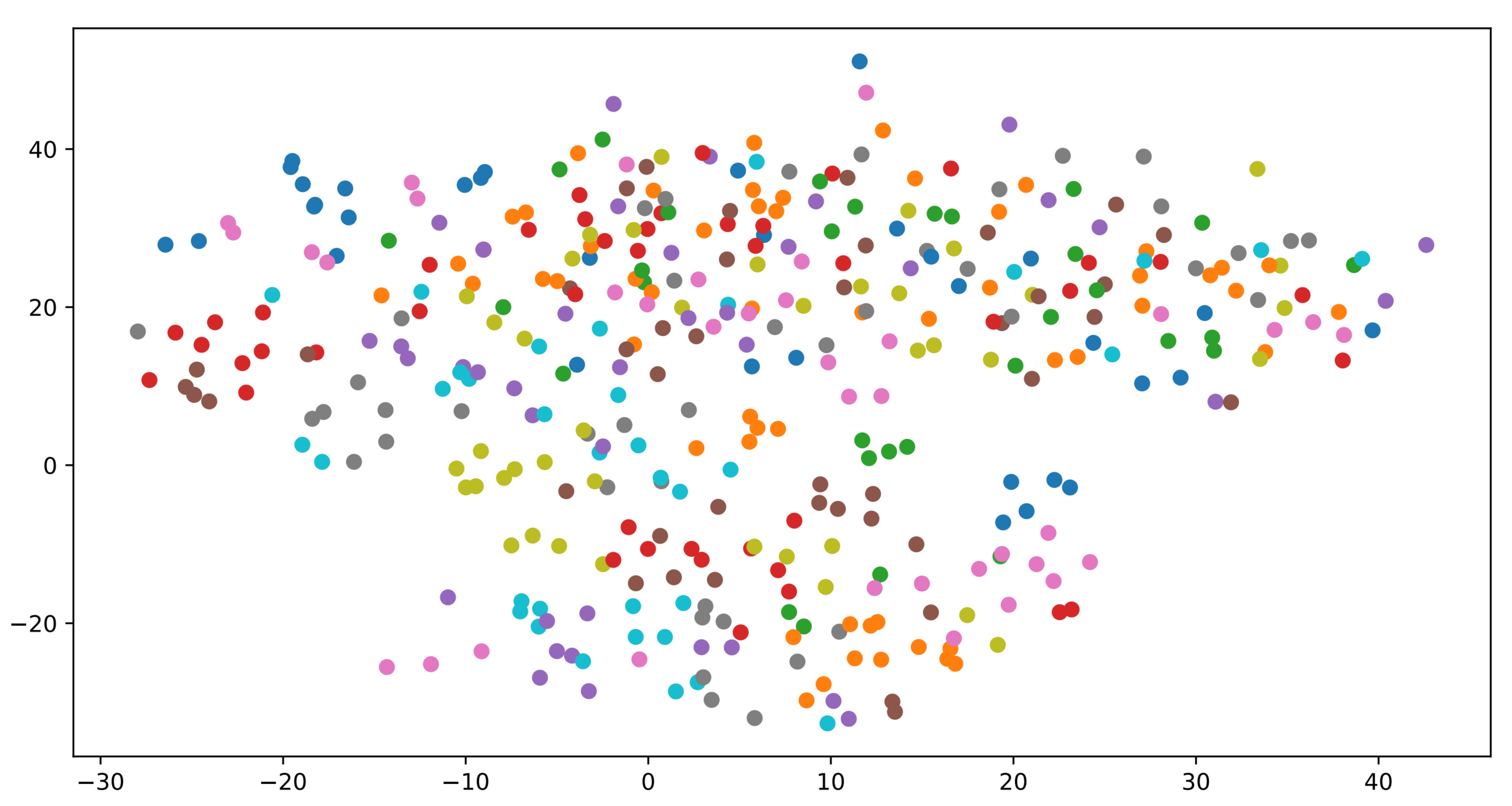}} \\
	 \rotatebox[origin=c]{90}{MTDA-ITA} & 
	 \subfloat[]{\label{fig:embed_d_all}\includegraphics[trim={3.2cm 2.3cm 0cm 0cm},clip,height = .16\linewidth,width=4.65cm,valign=c]{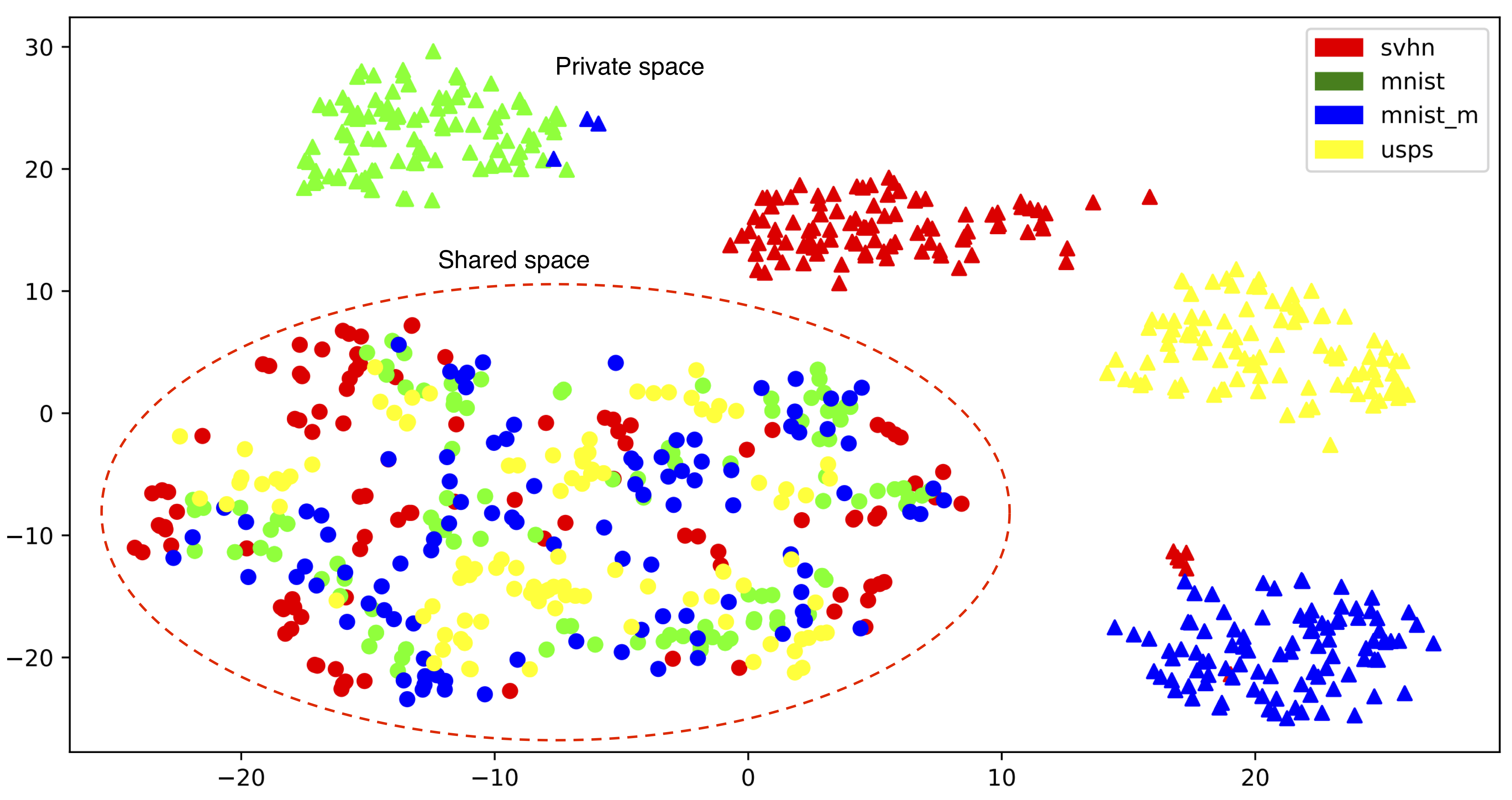}} & 
	\subfloat[]{\label{fig:embed_c_all}\includegraphics[trim={3.2cm 2.3cm 0cm 0cm},clip,height = .16\linewidth, width=4.65cm,valign=c]{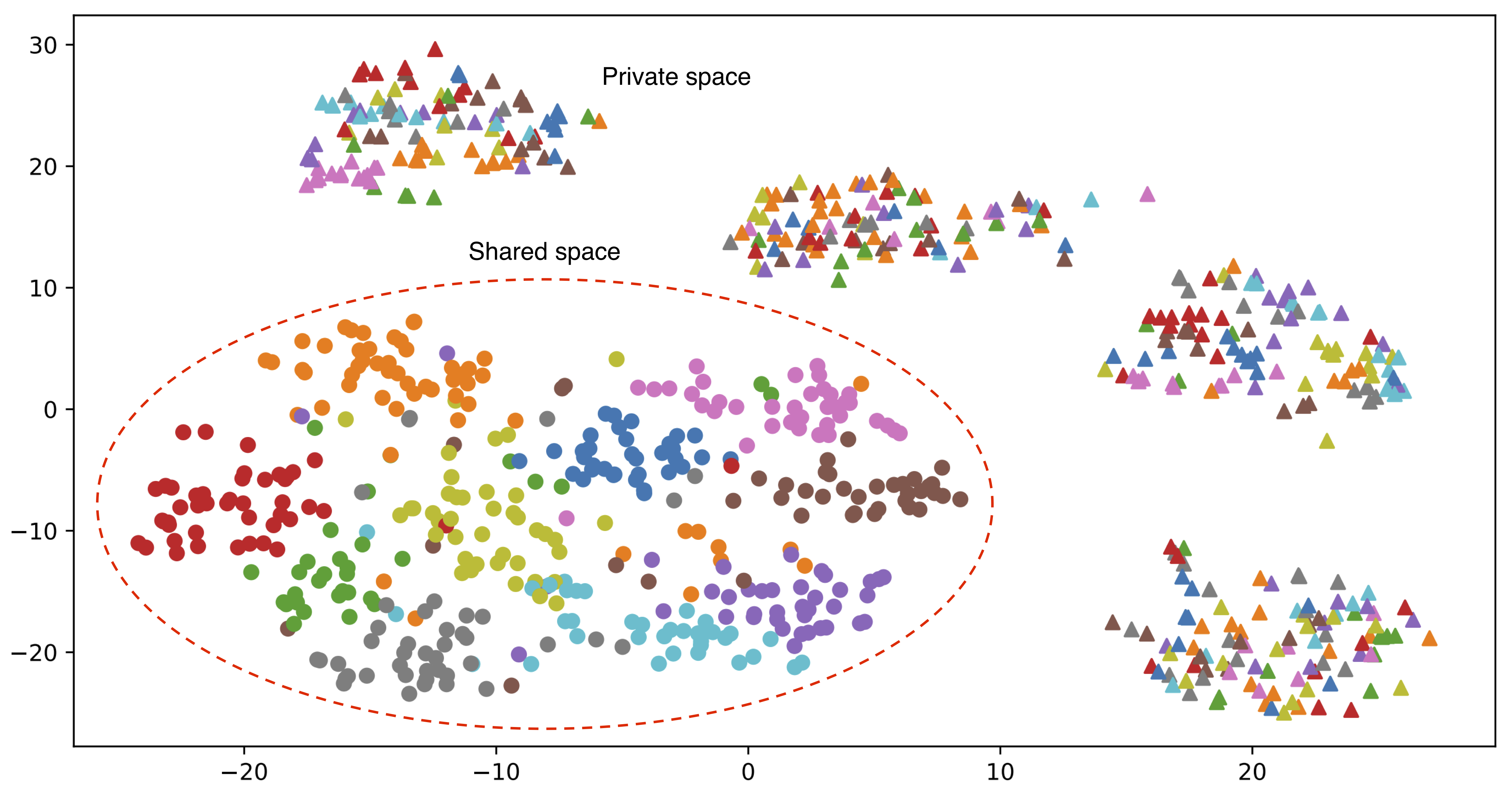}}\\
	\rotatebox[origin=c]{90}{MTDA-woE} & 
	\subfloat[]{\label{fig:embed_d_woE}\includegraphics[trim={1.38cm .93cm 1cm .8cm},clip,height = .16\linewidth, width=4.6cm,valign=c]{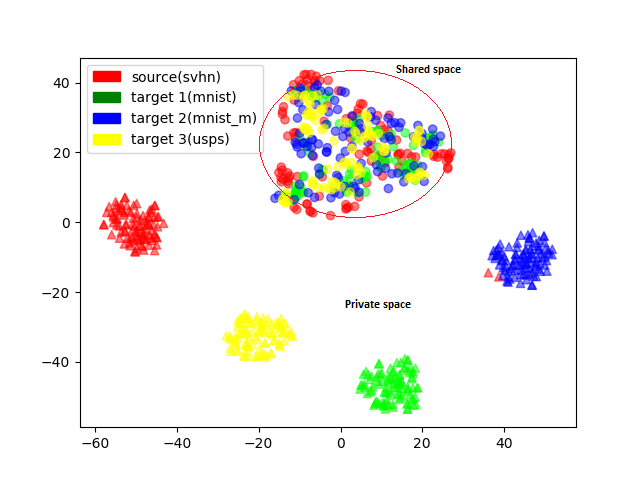}} &
	\subfloat[]{\label{fig:embed_c_woE}\includegraphics[trim={1.38cm .93cm 1cm .8cm},clip,height = .16\linewidth, width=4.6cm,valign=c]{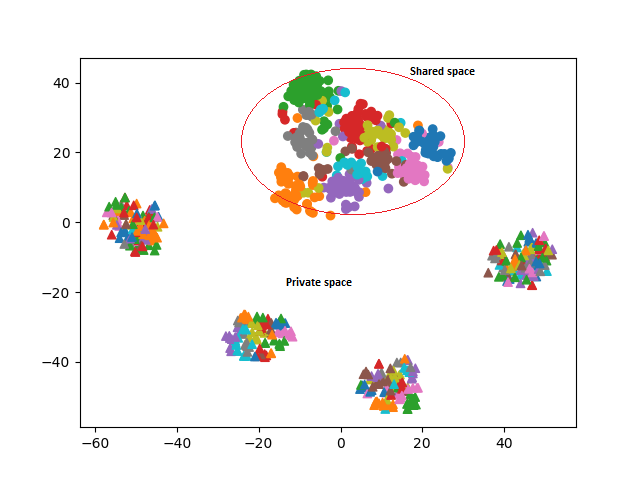}}\\
	\rotatebox[origin=c]{90}{MTDA-woP} & 
	\subfloat[]{\label{fig:embed_d_woP}\includegraphics[trim={1.38cm .94cm 1cm .8cm},clip,height = .16\linewidth, width=4.6cm,valign=c]{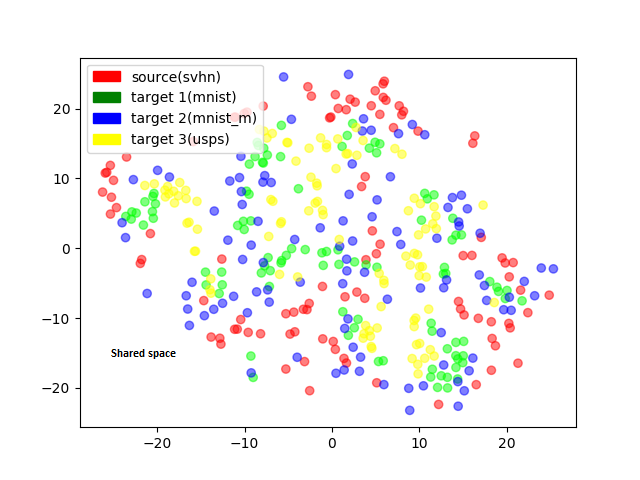}} & 
	\subfloat[]{\label{fig:embed_c_woP}\includegraphics[trim={1.38cm .94cm 1cm .8cm},clip,height = .16\linewidth, width=4.6cm,valign=c]{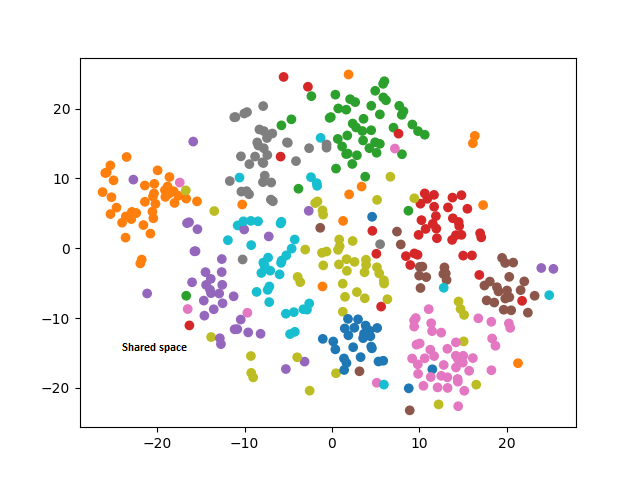}}\\
	\rotatebox[origin=c]{90}{MTDA-woR} & 
	\subfloat[]{\label{fig:embed_d_woR}\includegraphics[trim={1.38cm .94cm 1cm .8cm},clip,height = .16\linewidth, width=4.6cm,valign=c]{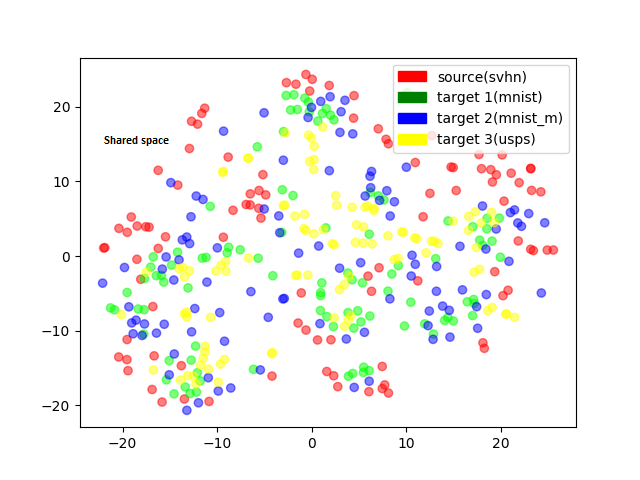}} &
	\subfloat[]{\label{fig:embed_c_woR}\includegraphics[trim={1.38cm .94cm 1cm .8cm},clip,height = .16\linewidth, width=4.6cm,valign=c]{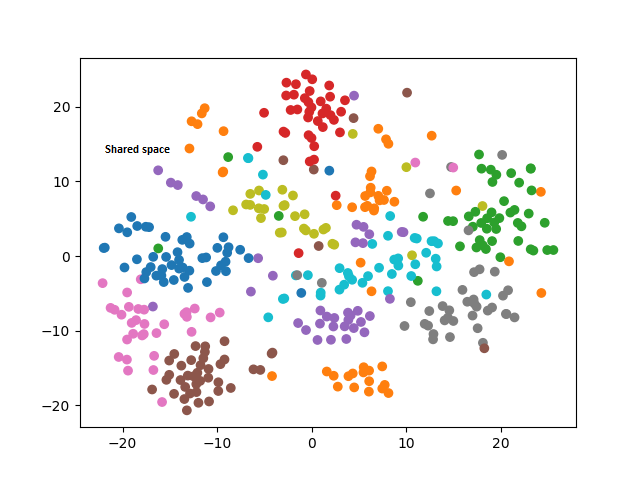}}\\
	\rotatebox[origin=c]{90}{MTDA-woD} & 
	\subfloat[]{\label{fig:embed_d_woD}\includegraphics[trim={1.38cm .94cm 1cm .8cm},clip,height = .16\linewidth, width=4.6cm,valign=c]{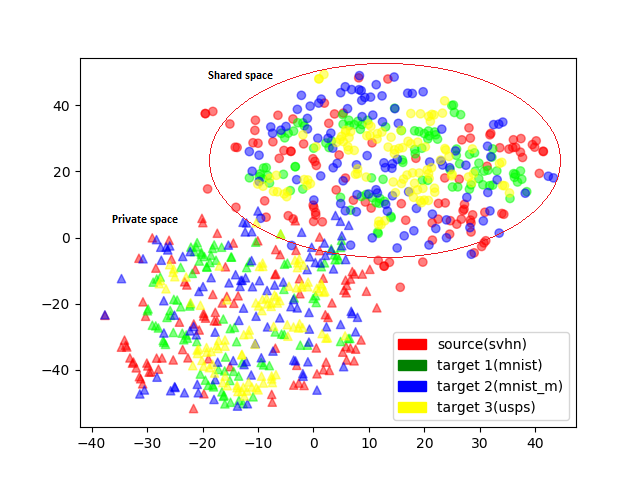}} &
	\subfloat[]{\label{fig:embed_c_woD}\includegraphics[trim={1.37cm .94cm 1cm .8cm},clip,height = .16\linewidth, width=4.6cm,valign=c]{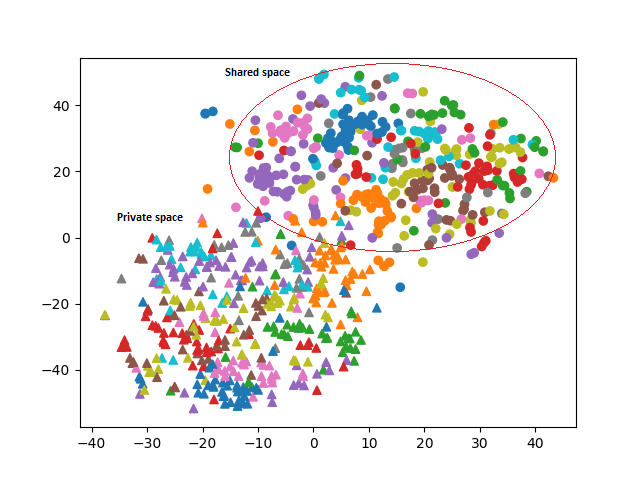}}
     \end{tabular}
     \caption{Feature visualization for embedding of digit datasets using t-SNE algorithm. The first and the second columns show the domains and classes, respectively, with color indicating domain and class membership. \protect\subref{fig:embed_d_orig},\protect\subref{fig:embed_c_orig} Original features.
     \protect\subref{fig:embed_d_all},\protect\subref{fig:embed_c_all} learned features for \textbf{MTDA-ITA} (triangle marker: private features, circle marker: shared features). Large clusters in the right column represent points from the shared space, while the smaller ones are from the private spaces. The remaining figures depict the learned features without: \protect\subref{fig:embed_d_woE},\protect\subref{fig:embed_c_woE} the classifier entropy loss, \textbf{MTDA-woE}; \protect\subref{fig:embed_d_woP},\protect\subref{fig:embed_c_woP} the private encoder, \textbf{MTDA-woP}; \protect\subref{fig:embed_d_woR},\protect\subref{fig:embed_c_woR} the reconstruction loss/decoder, \textbf{MTDA-woR}; and \protect\subref{fig:embed_d_woD},\protect\subref{fig:embed_c_woD} the multi-domain separation loss, \textbf{MTDA-woD}.
     }
     \label{vis}
\end{figure}

We use t-SNE~\cite{maaten2008visualizing} on Digit dataset to visualize
 shared and private feature representations from different domains. 
 Fig.~\ref{vis} shows shared and private features from source (\textbf{SVHN}) and target domains before \subref{fig:embed_d_orig},\subref{fig:embed_c_orig} and after
adaptation \subref{fig:embed_d_all},\subref{fig:embed_c_all}. \textbf{MTDA-ITA} significantly reduces the domain mismatch for the shared features (circle markers in \cref{fig:embed_c_all}, strong mixing of domain labels in this cluster, \cref{fig:embed_d_all}) and increases it for the private features (triangle markers, pure and well-separated domain clusters in \cref{fig:embed_d_all}). This is partially due to the proposed multi-domain separation loss through the use of the domain classifier $D$, which penalizes the domain mismatch for the shared features and rewards the mismatch for the private features. Moreover, as supported by the quantitative results in Tab.~\ref{1}, joint adaptation of related domains and the classifier, accomplished through the model, leads to superior class separability, compared to original features.  This is depicted in \cref{fig:embed_c_all}, where the points in the shared space (large cluster) are grouped into class-specific subgroups (color indicates class label), while they are mixed in private spaces (smaller clusters).  This is in contrast to \cref{fig:embed_c_orig}, where original features show no class-specificity. 

We also show the learned shared and private features for the models \textbf{MTDA-woE}, \textbf{MTDA-woP}, \textbf{MTDA-woR}, and \textbf{MTDA-woD},  in \crefrange{fig:embed_d_woE}{fig:embed_c_woD}. Note that since the private encoder $E_p$ is disabled for \textbf{MTDA-woR}, and \textbf{MTDA-woP}, no private features are depicted in \crefrange{fig:embed_d_woP}{fig:embed_c_woR}. %Fig.~\ref{vis}(i-l). 
The class label separation in the shared space for \textbf{MTDA-woE}, \textbf{MTDA-woP}, and \textbf{MTDA-woR}, \cref{fig:embed_c_woE,fig:embed_c_woP,fig:embed_c_woR}, is still evident but not as strong as in the full model, \cref{fig:embed_c_all}, corroborating the small loss in classification accuracy observed in \cref{fig:abl_svhn}.  On the other hand, \textbf{MTDA-woD} has significant mixing of class labels in the shared space, \cref{fig:embed_c_woD}, more so than \textbf{MTDA-woE}, \textbf{MTDA-woR}, and \textbf{MTDA-woP}, implying worse classification prediction in \cref{fig:abl_svhn} due to the severe mismatch between different domains. 
%from which we can see that with all modules except the domain classifier $D$, the model is not able to distinguish between the shared and the private features of the different domains, which results in the performance degradation due to the severe domain mismatch between domains (Our ablation study in Fig.~\ref{ablation} is relevant to supporting this claim.) 
\section{Conclusion}
This paper presented an information theoretic end-to-end approach to \textbf{uDA} in the context of a single source and multiple target domains that share a common task or properties. The proposed method learns feature representations invariant under multiple domain shifts and simultaneously discriminative for the learning task. This is accomplished by explicitly separating representations private to each domain and shared between source and target domains using a novel discrimination strategy. Our use of a single private domain encoder results in a highly scalable model, easily optimized using established back-propagation approaches. Results on three benchmark datasets for image classification show superiority of the proposed method compared to the state-of-the-art methods for unsupervised domain adaptation of visual domain categories.
\bibliographystyle{abbrvnat}
\bibliography{egbib}

%\newpage

\begin{appendix}
\section{Network Architecture}
\begin{table}[tbh!]
\caption{\small Network architecture for the experiments.}
\label{tbl::arch}
\small
\centering
\begin{tabularx}{265pt}{clc}
\toprule
Layer &  Encoders (shared, private)  \tabularnewline\midrule
1 & CONV-(N16,K7,S1), \;ReLU \tabularnewline
2 & CONV-(N32,K3,S2), \;ReLU \tabularnewline
3 & CONV-(N64,K3,S2), \;ReLU  \tabularnewline
4 & RESBLK-(N64,K3,S1) \tabularnewline
5 & RESBLK-(N64,K3,S1) \tabularnewline
6 & RESBLK-(N64,K3,S1) \tabularnewline
7 & RESBLK-(N64,K3,S1) \tabularnewline\midrule
Layer &  Decoder\tabularnewline\midrule
1 & RESBLK-(N64,K3,S1)\tabularnewline
2 & RESBLK-(N64,K3,S1) \tabularnewline
3 & RESBLK-(N64,K3,S1) \tabularnewline
4 & RESBLK-(N64,K3,S1) \tabularnewline
5 &  DCONV-(N32,K3,S2), \;ReLU \tabularnewline
6 &  DCONV-(N16,K3,S2), \;ReLU \tabularnewline
7 &  DCONV-(N1,K1,S1), \;TanH \tabularnewline\midrule
Layer &  Discriminator  \tabularnewline\midrule
1 & CONV-(N4,K3,S1), ReLU \tabularnewline
2 & CONV-(N8,K3,S1), ReLU  \tabularnewline
3 & CONV-(N16,K3,S1), ReLU  \tabularnewline
4 & CONV-(N32,K3,S1), ReLU  \tabularnewline
5 & CONV-(N1,K3,S1), ReLU  \tabularnewline
6 & DENSE-(ND), Softmax \tabularnewline \midrule
Layer &  Classifier  \tabularnewline\midrule
1 & CONV-(N4,K3,S1), ReLU \tabularnewline
2 & CONV-(N8,K3,S1), ReLU  \tabularnewline
3 & CONV-(N16,K3,S1), ReLU  \tabularnewline
4 & CONV-(N32,K3,S1), ReLU  \tabularnewline
5 & CONV-(N1,K3,S1), ReLU  \tabularnewline
6 & DENSE-(NC), Softmax \tabularnewline\bottomrule

\end{tabularx}
\end{table}

The network architecture used for the experiments is given in Table~\ref{tbl::arch}. We use the following abbreviation for ease of presentation: N=Neurons, K=Kernel size, S=Stride size, D=Number of Domains, C=number of Classes. The transposed convolutional layer is denoted by DCONV. The residual basic block is denoted as RESBLK. 
\end{appendix}

\end{document}